%% file: 0_main.tex
\documentclass{article}

\PassOptionsToPackage{numbers, compress}{natbib}
\usepackage[main,final]{neurips_2026}

\usepackage[dvipsnames]{xcolor}
\usepackage[most]{tcolorbox}
\usepackage{graphicx}
\usepackage{tabularx}

\lstdefinestyle{jsonstyle}{
  basicstyle=\ttfamily\small,
  columns=fullflexible,
  keepspaces=true,
  showstringspaces=false,
  breaklines=true,
  frame=none,
  aboveskip=0pt,
  belowskip=0pt
}

\usepackage{xcolor}
\usepackage{soul}

\usepackage{enumitem}
\usepackage[utf8]{inputenc} 
\usepackage[T1]{fontenc}    
\usepackage{hyperref}       
\usepackage{url}            
\usepackage{booktabs}       
\usepackage{amsfonts}       
\usepackage{nicefrac}       
\usepackage{microtype}      
\usepackage{xcolor}         
\usepackage{amsmath}
\usepackage{wrapfig}

\usepackage{multirow}
\usepackage[table,xcdraw]{xcolor}

\newcommand{\eg}{\emph{e.g., }}

\title{UniCustom: Unified Visual Conditioning for Multi-Reference Image Generation}

\author{%
  Yiyan Xu$^{1}$\footnote[1]\ \ , 
  Qiulin Wang$^{2}$\footnote[2]\ \ \footnote[3]\ \ ,
  Wenjie Wang$^{1}$\footnote[2]\ \ ,
  Yunyao Mao$^{2}$, \ \ \\
  \textbf{Xintao Wang}$^{2}$, \ \
  \textbf{Pengfei Wan}$^{2}$, \ \
  \textbf{Kun Gai}$^{2}$, \ \
  \textbf{Fuli Feng}$^{1}$\ \ \\
  \small $^1$University of Science and Technology of China, 
    $^2$KlingAI Research \\
  {\tt\small yiyanxu24@gmail.com, qiulin\_wang@foxmail.com, wenjiewang96@gmail.com} \\
  {\small \textcolor{magenta}{\url{https://yiyanxu.github.io/UniCustom/}}}
}

\begin{document}

\renewcommand{\thefootnote}{\fnsymbol{footnote}} 
\footnotetext[1]{This work was conducted during the author's internship at KlingAI.}
\footnotetext[2]{Corresponding authors.}
\footnotetext[3]{Project Lead.}
\renewcommand{\thefootnote}{\arabic{footnote}} 

\maketitle

\begin{abstract}
    Multi-reference image generation aims to synthesize images from textual instructions while faithfully preserving subject identities from multiple reference images. Existing VLM-enhanced diffusion models commonly rely on decoupled visual conditioning: semantic ViT features are processed by the VLM for instruction understanding, whereas appearance-rich VAE features are injected later into the diffusion backbone. Despite its intuitive design, this separation makes it difficult for the model to associate each semantically grounded subject with visual details from the correct reference image. As a result, the model may recognize which subject is being referred to, but fail to preserve its identity and fine-grained appearance, leading to attribute leakage and cross-reference confusion in complex multi-reference settings. To address this issue, we propose \textbf{\textit{UniCustom}}, a unified visual conditioning framework that fuses ViT and VAE features \textit{before} VLM encoding. This early fusion exposes the VLM to both semantic cues and appearance-rich details, enabling its hidden states to jointly encode the referred subject and corresponding visual appearance with only a lightweight linear fusion layer. To learn such unified representations, we adopt a two-stage training strategy: reconstruction-oriented pretraining that preserves reference-specific appearance details in the fused hidden states, followed by supervised finetuning on single- and multi-reference generation tasks. We further introduce a slot-wise binding regularization that encourages each image slot to preserve low-level details of its corresponding reference, thereby reducing cross-reference entanglement. Experiments on two multi-reference generation benchmarks demonstrate that UniCustom consistently improves subject consistency, instruction following, and compositional fidelity over strong baselines.
\end{abstract}

\input{1_intro}
\input{2_method}
\input{3_exps}

\input{4_related_work}

\input{5_conclusion}

\input{acknowledgement}

\bibliographystyle{plainnat}
\bibliography{bibfile}

\appendix

\input{appendix}




\end{document}

%% file: 1_intro.tex
\section{Introduction}

Recent advances in text-to-image generation have substantially improved the fidelity, diversity, and controllability of visual synthesis~\cite{gpt-image-2, nanobanan, cai2025z, li2026easier, wang2025precise, li2026speed, qiu2025accelerating1, qiu2025accelerating2}. However, text alone is often insufficient for specifying fine-grained visual intent in practical creation scenarios. Users may wish to preserve and recombine concrete subjects from reference images, such as a particular person, object, garment, scene, or style. This has motivated increasing interest in \textit{multi-reference image generation}, where a model is given multiple reference images and a textual instruction, and is expected to synthesize a coherent image that follows the instruction and preserves the specified subject identities and appearances from the references~\cite{wu2025less,she2026mosaic,chen2026xverse}.

Multi-reference generation poses a unique challenge beyond standard text-to-image synthesis, where the textual instruction not only describes the target scene, but also specifies how subjects from different references should be selected, composed, and rendered~\cite{wang2025scone,garibi2025tokenverse}. For example, an instruction may require ``the woman from Picture 1'' to wear ``the hat from Picture 2'' while interacting with ``the dog from Picture 3''. Successful generation therefore requires two essential capabilities: 1) \textbf{\textit{Semantic grounding}}, identifying which reference image, subject, or region is being referred to by each textual expression~\cite{chen2026xverse,garibi2025tokenverse}; 2) \textbf{\textit{Visual binding}}, associating each grounded subject with its corresponding appearance, identity, texture, and fine-grained attributes throughout the generation process~\cite{mou2025dreamo}. These two capabilities are related but not equivalent. A model may correctly understand which subject is requested, yet still render that subject with attributes from another reference, leading to identity confusion, attribute leakage, missing entities, or incorrect compositions.

\begin{wrapfigure}{r}{0.45\textwidth}
\vspace{-1em}
\centering
\includegraphics[width=0.45\textwidth]{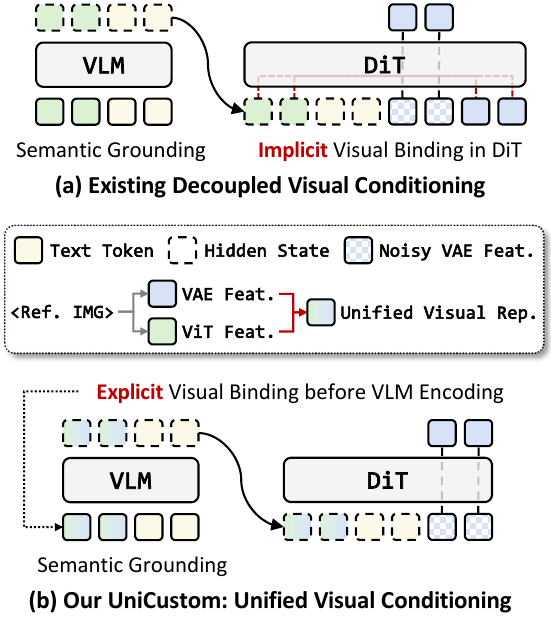}
\vspace{-1.5em}
\caption{\textbf{Illustration of decoupled and unified visual conditioning.}}
\vspace{-1.2em}
\label{fig:intro}
\end{wrapfigure}

Recent VLM-enhanced diffusion models (\eg OmniGen2~\cite{wu2025omnigen2}, Qwen-Image-Edit~\cite{wu2025qwen}, LongCat-Image-Edit~\cite{team2025longcat}) provide a promising framework for this task by leveraging the multimodal understanding and instruction-following ability of Vision-Language Models (VLMs). As illustrated in Figure~\ref{fig:intro}(a), a common design encodes reference images through two separate visual pathways. High-level ViT features are fed into the VLM together with the textual instruction, providing semantics for instruction understanding and semantic grounding. In parallel, VAE features, which preserve low-level visual details, are injected into the Diffusion Transformer (DiT) directly during generation to support faithful visual synthesis. This design is intuitive and has been widely adopted in reference-based generation~\cite{deng2025cinema,hu2025hunyuancustom}.

However, we argue that such decoupled design introduces a fundamental \textit{grounding--binding gap}. Since the VLM only accesses semantic ViT features, its hidden states are well-suited for semantic grounding and instruction-level reasoning, but lack fine-grained appearance cues required for faithful rendering~\cite{deng2025cinema}. In contrast, VAE features, which preserve such appearance details, are injected only later into the DiT. Consequently, even when the VLM correctly grounds an instruction to the intended subjects, the DiT must still infer how these later-injected visual details should be associated with the VLM-encoded hidden states during generation, as illustrated in Figure~\ref{fig:intro}(a). Such implicit visual binding becomes unreliable in multi-reference scenarios, where multiple subjects with similar appearances may coexist across references. The model may therefore follow the instruction at the semantic level while binding visual details to the wrong subject.

To address this issue, we propose \textbf{\textit{UniCustom}}, a unified visual conditioning framework for multi-reference image generation. As shown in Figure~\ref{fig:intro}(b), UniCustom bridges the grounding--binding gap by fusing ViT and VAE features \textit{before} VLM encoding. The resulting unified visual representation integrates ViT-derived semantic cues for reference grounding with VAE-derived appearance cues for faithful rendering. When jointly encoded with the textual instruction, it enables the VLM to produce hidden states that are both semantically addressable and appearance-aware, thereby providing the DiT with explicit semantic-visual correspondences during generation. Notably, our design is simple and lightweight: only a single linear fusion layer is sufficient to merge the two feature spaces, yielding substantial improvements in multi-reference grounding and visual consistency.

To make the unified visual representation effective for generation, we adopt a two-stage training strategy. In the pretraining stage, the model is optimized primarily on reconstruction-oriented tasks to establish semantic grounding, visual binding, and alignment between the unified visual representation and the DiT. During supervised fine-tuning, the model is mainly trained on single- and multi-reference image generation tasks, enabling the DiT to exploit VLM hidden states derived from the unified visual representation for reference-based synthesis.
To further provide more structured conditioning signals for the DiT, we introduce slot-wise binding regularization on the VLM hidden states. This encourages each image slot to preserve reference-specific visual details while reducing cross-reference entanglement, allowing the DiT to parse and utilize multi-reference information more effectively.

To summarize, our contributions are as follows:
\begin{itemize}[leftmargin=*]
    \item We identify the \textit{grounding--binding gap} in VLM-enhanced diffusion models for multi-reference image generation. In existing decoupled conditioning designs, the DiT must implicitly associate VLM-encoded subject semantics with separately injected appearance features, which becomes unreliable with multiple references.

    \item We propose \textbf{\textit{UniCustom}}, a unified visual conditioning framework that makes reference appearances semantically accessible. By fusing ViT and VAE features before VLM encoding, UniCustom produces hidden states that jointly encode the referred subject and its fine-grained visual details, thereby providing the DiT with more explicit semantic--appearance correspondences.

    \item We introduce a two-stage training strategy with slot-wise binding regularization to progressively learn reference-specific appearance preservation and adapt it to multi-reference generation. The reconstruction-oriented pretraining stage achieves a single-image reconstruction PSNR close to $30$ dB, indicating that fused VLM hidden states can serve as an effective conduit for transmitting low-level details from VAE features to the DiT. Extensive experiments on two multi-reference image generation benchmarks further demonstrate that UniCustom outperforms existing methods.
\end{itemize}

%% file: 2_method.tex
\section{Method}

\subsection{Model Architecture}
\begin{figure}[t]
    \centering
    \setlength{\abovecaptionskip}{0.2cm}
    \setlength{\belowcaptionskip}{-0.2cm}
    \includegraphics[width=0.9\linewidth]{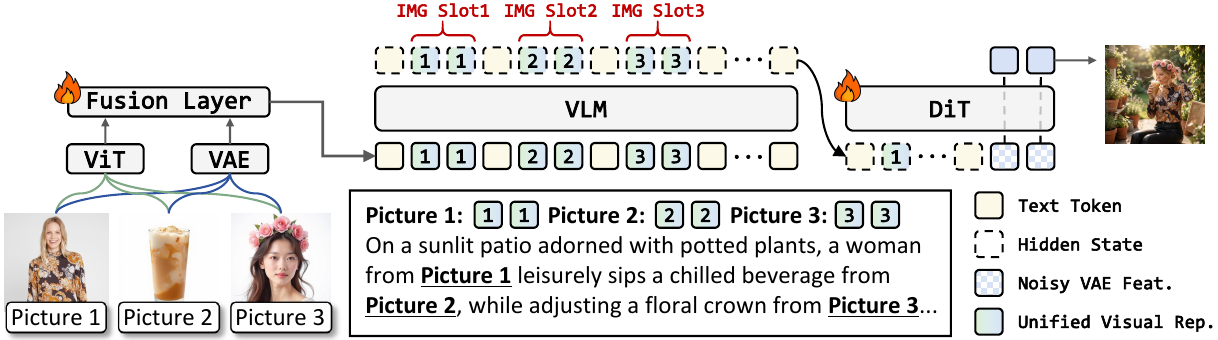}
    \caption{\textbf{Overview of UniCustom}. UniCustom fuses ViT and VAE features \textit{before} VLM encoding, producing semantically addressable and appearance-aware hidden states for DiT generation.}
    \label{fig:overview}
\end{figure}

As shown in Figure~\ref{fig:overview}, our model builds upon VLM-enhanced diffusion models, in which the VLM encodes textual instructions and reference images into hidden states that serve as conditioning signals for the DiT during image generation. 
Our key architectural modification is a lightweight early-fusion module that injects VAE features into ViT features before VLM encoding. This design enables the VLM hidden states to incorporate both semantic grounding cues and fine-grained appearance details, thereby providing the DiT with more informative and reference-aware conditioning for generation.

\paragraph{Unified visual representation via early fusion.}
For each reference image, we extract a sequence of ViT features and VAE features, denoted as $\mathbf{F}^{\mathrm{vit}} \in \mathbb{R}^{L \times d_{\mathrm{vit}}}$ and $\mathbf{F}^{\mathrm{vae}} \in \mathbb{R}^{L \times d_{\mathrm{vae}}}$, respectively, where $L$ represents the sequence length. Note that each reference image is resized so that the resulting ViT and VAE feature sequences have the same length and are spatially aligned.
We concatenate the two feature sequences along the channel dimension and project them back to the ViT feature dimension using a lightweight linear fusion layer:
\begin{equation}
    \mathbf{F}^{\mathrm{uni}}=[\mathbf{F}^{\mathrm{vit}};\mathbf{F}^{\mathrm{vae}}]\mathbf{W}_{\mathrm{fuse}}+\mathbf{b}_{\mathrm{fuse}},
\end{equation}
where
$\mathbf{W}_{\mathrm{fuse}}\in\mathbb{R}^{(d_{\mathrm{vit}}+d_{\mathrm{vae}})\times d_{\mathrm{vit}}}$
and
$\mathbf{b}_{\mathrm{fuse}}\in\mathbb{R}^{d_{\mathrm{vit}}}$.
The resulting unified representation
$\mathbf{F}^{\mathrm{uni}}\in\mathbb{R}^{L\times d_{\mathrm{vit}}}$
has the same dimensionality as the original ViT features and can therefore be directly consumed by the pretrained VLM.

To preserve compatibility with the pretrained VLM at initialization, we adopt an identity-preserving initialization for the fusion layer. Specifically, the weights corresponding to the ViT feature dimensions are initialized as an identity mapping, while those corresponding to the VAE feature dimensions are initialized to zero:
\begin{equation}
    \mathbf{W}_{\mathrm{fuse}} = \begin{bmatrix}
        \mathbf{I}_{d_{\mathrm{vit}}} \\
        \mathbf{0}_{d_{\mathrm{vae}}\times d_{\mathrm{vit}}}
    \end{bmatrix},
    \quad
    \mathbf{b}_{\mathrm{fuse}} = \mathbf{0}_{d_{\mathrm{vit}}}.
\end{equation}
Under this initialization, $\mathbf{F}^{\mathrm{uni}}$ is initially equivalent to the original ViT feature. The model thus starts from the pretrained VLM-compatible visual feature space and gradually learns to incorporate fine-grained VAE appearance cues during training.

\paragraph{Multimodal encoding for DiT conditioning.}
Before VLM encoding, we organize the multimodal input sequence by interleaving explicit image identifiers with their corresponding unified visual representations. For $N$ reference images, the input sequence is formatted as the following example:
\begin{center}
\begin{tcolorbox}[
                  breakable,
                  colbacktitle=gray!30,
                  coltitle=black,
                  colback=white,
                  colframe=black,
                  width=\linewidth,
                  arc=1mm, auto outer arc,
                  boxrule=0.5pt,
                  left=3pt,
                  right=3pt,
                  top=1pt,
                  bottom=1pt,
                  middle=1pt,
                  halign title=left,
                  toprule=0.5pt,
                  bottomrule=0.5pt,
                  before upper={\setlength{\parskip}{1pt}},
                  before lower={\setlength{\parskip}{1pt}},
                 ]
Picture 1: <$\mathbf{F}_1^{\mathrm{uni}}$> Picture 2: <$\mathbf{F}_2^{\mathrm{uni}}$> \dots Picture N: <$\mathbf{F}_N^{\mathrm{uni}}$> The woman from Picture 1 wears the hat from Picture 2\dots walking along a river with the dog from Picture N.

\end{tcolorbox}
\end{center}
The explicit image identifiers act as language-level anchors for reference grounding, allowing textual expressions such as ``the woman from Picture 1'' or ``the hat from Picture 2'' to be associated with the corresponding visual representations. Since each visual representation has already integrated ViT semantics and VAE appearance cues, the VLM hidden states are both semantically addressable and appearance-aware, which serve as conditioning signals for the DiT, enabling the denoising process to rely on more explicit semantic-visual correspondences rather than inferring such bindings implicitly.

\paragraph{Positional encoding.}
For DiT conditioning, we adopt the Multimodal Rotary Position Embedding (MRoPE) from Qwen2.5-VL~\cite{bai2025qwen25vl}, which decomposes positional information into three components corresponding to the temporal, height, and width axes. For text positions, the same position index is assigned to all three components, making it equivalent to standard one-dimensional RoPE. For image positions, the temporal index is kept constant within each image, while the height and width indices are assigned according to the spatial location of each token. 

\subsection{Training Strategy}
To effectively learn the proposed unified visual representation and align it with the diffusion denoising process, we adopt a two-stage training strategy. Throughout training, the VLM is kept frozen, and only the lightweight feature fusion layer and the DiT are optimized. This design preserves the pretrained multimodal understanding ability of the VLM, while allowing the model to progressively learn how to inject fine-grained visual details into the VLM hidden states and adapt the diffusion backbone to the resulting conditioning signals. Details about the training data can be found in Appendix~\ref{sec:data}.

\begin{figure}[t]
    \centering
    \setlength{\abovecaptionskip}{0.2cm}
    \setlength{\belowcaptionskip}{-0.2cm}
    \includegraphics[width=\linewidth]{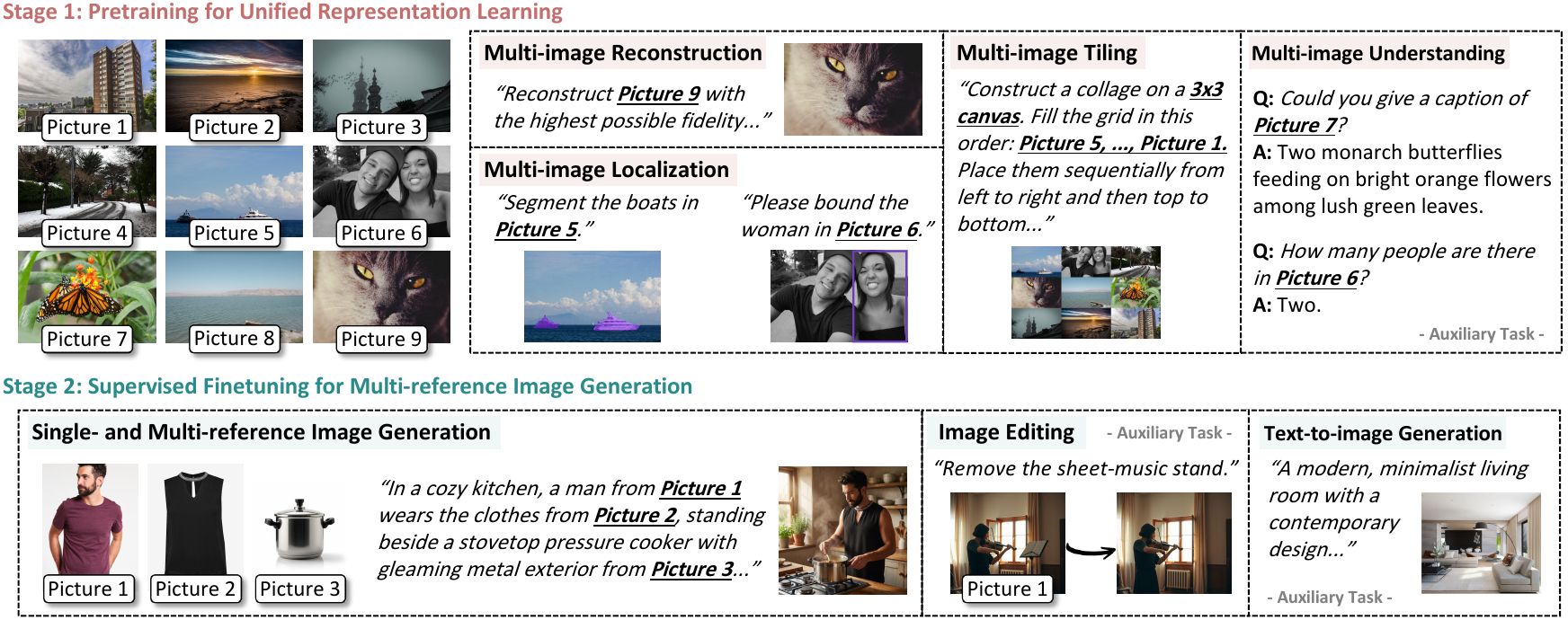}
    \caption{\textbf{Two-stage training strategy.} The first stage progressively learns a unified visual representation that supports fine-grained reference encoding, semantic grounding, and reliable textual-to-visual binding through reconstruction-oriented multi-image pretraining. The second stage further adapts the diffusion backbone to reference-based image generation, enabling instruction-following synthesis with single or multiple reference images while preserving the learned grounding and binding abilities.}
    \label{fig:training}
\end{figure}

\paragraph{Stage 1: Pretraining for unified representation learning.}
In the first stage, we jointly optimize the fusion layer and the DiT with reconstruction-oriented tasks, aiming to learn unified visual representations that support reliable semantic grounding and visual binding, and adapt the DiT to interpret the resulting VLM hidden states during denoising.
As illustrated in Figure~\ref{fig:training}, the pretraining tasks include multi-image reconstruction, localization, and tiling. These reconstruction-oriented tasks require the model to identify the specified reference according to textual instructions and generate the corresponding visual output. In doing so, they encourage the VLM hidden states to encode reference semantics and fine-grained appearance cues, while simultaneously optimizing the DiT to parse these hidden states into the corresponding visual outputs.

To prevent the unified representation from being dominated by VAE-derived low-level visual details, we further mix a small proportion of multi-image understanding tasks into pretraining. Since the VLM is kept frozen throughout training, these tasks primarily regularize the fusion layer, encouraging the unified representation to remain compatible with the semantic structure expected by the VLM. After this stage, the model can transmit fine-grained appearance information from the VAE through the frozen VLM, maintain reliable textual-to-visual reference binding, and provide hidden states that are readily usable by the DiT denoising process.

\paragraph{Stage 2: Supervised finetuning for multi-reference image generation.}
In the second stage, we conduct Supervised Finetuning (SFT) on diverse single- and multi-reference image generation tasks, while freezing the fusion layer and updating only the DiT. This stage adapts the pretrained hidden states from reconstruction-oriented learning to realistic reference-based generation, enabling the DiT to synthesize instruction-following outputs that preserve the specified reference appearances. In addition, we mix in a small amount of text-to-image and image editing data to improve general instruction adherence, and retain a small portion of the pretraining tasks to mitigate forgetting of the reference grounding and binding ability acquired in the first stage.

\paragraph{Slot-wise Binding Regularization.}
When unified visual representations from multiple input images are jointly encoded with the text instruction, the resulting hidden states may contain entangled visual information across references (see empirical analysis in Section~\ref{sec:regularization}). This is partly due to the decoder-only formulation adopted by modern VLMs, where visual and textual tokens are processed under causal attention. Consequently, each token is contextualized by its preceding tokens. In multi-image inputs, the hidden states corresponding to a later image may therefore incorporate information from earlier references. While such contextualization is useful for integrating visual evidence with the textual instruction, it also blurs the localization of VAE-level visual details within their original positions, which makes the final hidden states less structured and harder for the DiT to parse.

To obtain more structured VLM hidden states, we introduce a slot-wise binding regularization during pretraining. As depicted in Figure~\ref{fig:overview}, we define an image slot as the group of hidden states at the image-token positions assigned to a specific input image in the VLM sequence. Ideally, each slot should retain the VAE-level visual details of its corresponding image in a localized and decodable form. Specifically, for the $i$-th image, we take the hidden states $\mathbf{H}_i$ at its image slot and map them back to the VAE latent space using a single-layer projector $P(\cdot)$. The projected features are then supervised by the original VAE feature $\mathbf{F}_i^{\mathrm{vae}}$ with a mean squared error loss:
\begin{equation}
    \mathcal{L}_{\mathrm{bind}}=\dfrac{1}{N}\sum_{i=1}^N\left \| P(\mathbf{H}_i)-\mathbf{F}_i^{\mathrm{vae}} \right \|_2^2.
\end{equation}
This auxiliary objective encourages each image’s visual details to remain recoverable from its own slot, making the final VLM hidden states more explicitly organized and easier for the DiT to parse. The projector is used only during pretraining and discarded afterwards.

%% file: 3_exps.tex
\section{Experiments}

\input{tables/omnicontext}

\begin{figure}[t]
    \centering
    \setlength{\abovecaptionskip}{0.cm}
    \setlength{\belowcaptionskip}{-0.2cm}
    \includegraphics[width=\linewidth]{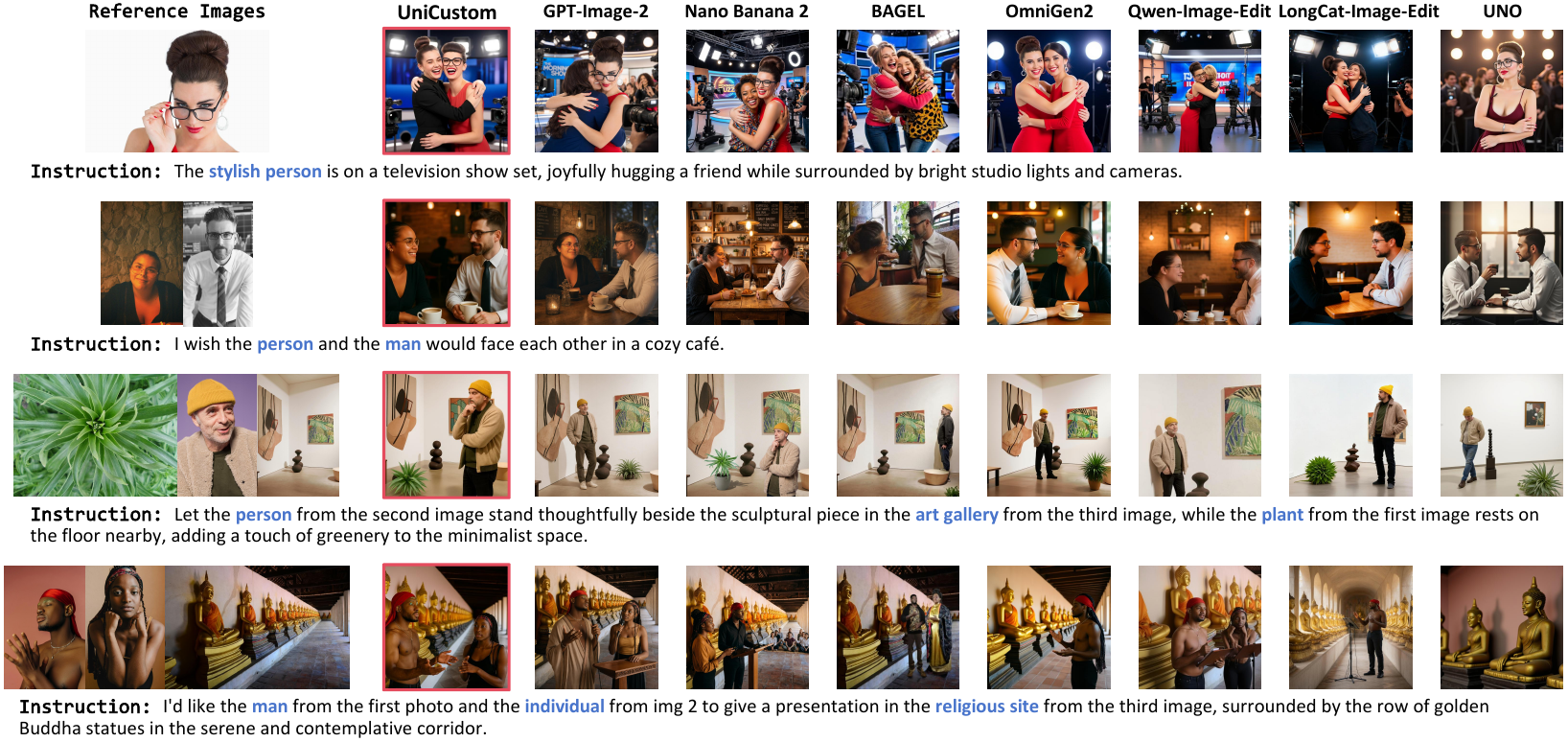}
    \caption{\textbf{Qualitative comparison on OmniContext~\cite{wu2025omnigen2}.}}
    \label{fig:case_omnicontext}
\end{figure}

\subsection{Implementation Details}
\label{sec:implementation}
We adopt Qwen2.5-VL~\cite{bai2025qwen25vl} as the VLM backbone and initialize the DiT from LongCat-Image-Edit~\cite{team2025longcat}. The VLM is kept frozen throughout training. Unless otherwise specified, all training samples are processed at resolutions no larger than $512\times512$. 
For evaluation, we report results on OmniContext~\cite{wu2025omnigen2} and MICo-Bench~\cite{wei2025mico}. For our model and open-source baselines, we sample images at $512\times512$ resolution using the default inference configuration of each method, with a fixed random seed. For closed-source models, we use their native output resolution of $1024\times1024$, as they do not support direct generation at $512\times512$.
More details can be found in Appendix~\ref{sec:implementation_appendix}.

\input{tables/micobench}

\subsection{Main Results}

\paragraph{Quantitative evaluation.}
Overall, UniCustom achieves the best performance among open-source models on both OmniContext~\cite{wu2025omnigen2} and MICo-Bench~\cite{wei2025mico}. As shown in Tables~\ref{tab:omnicontext} and~\ref{tab:micobench}, the gains are especially pronounced in multi-reference, scene-level, Object, HOI, and De\&Re settings, which require accurate reference grounding, subject-level appearance preservation, and compositional reasoning across multiple inputs. These results verify that fusing ViT and VAE features before VLM encoding, combined with the proposed two-stage training strategy, is effective for improving semantic grounding and visual binding in multi-image reference tasks. As a result, UniCustom can better model complex relationships among referenced subjects and maintain subject consistency under diverse generation scenarios. Despite the remaining gap to closed-source models, UniCustom sets a strong open-source baseline on both benchmarks, demonstrating the promise of the proposed unified visual conditioning.

\begin{figure}[t]
    \centering
    \includegraphics[width=\linewidth]{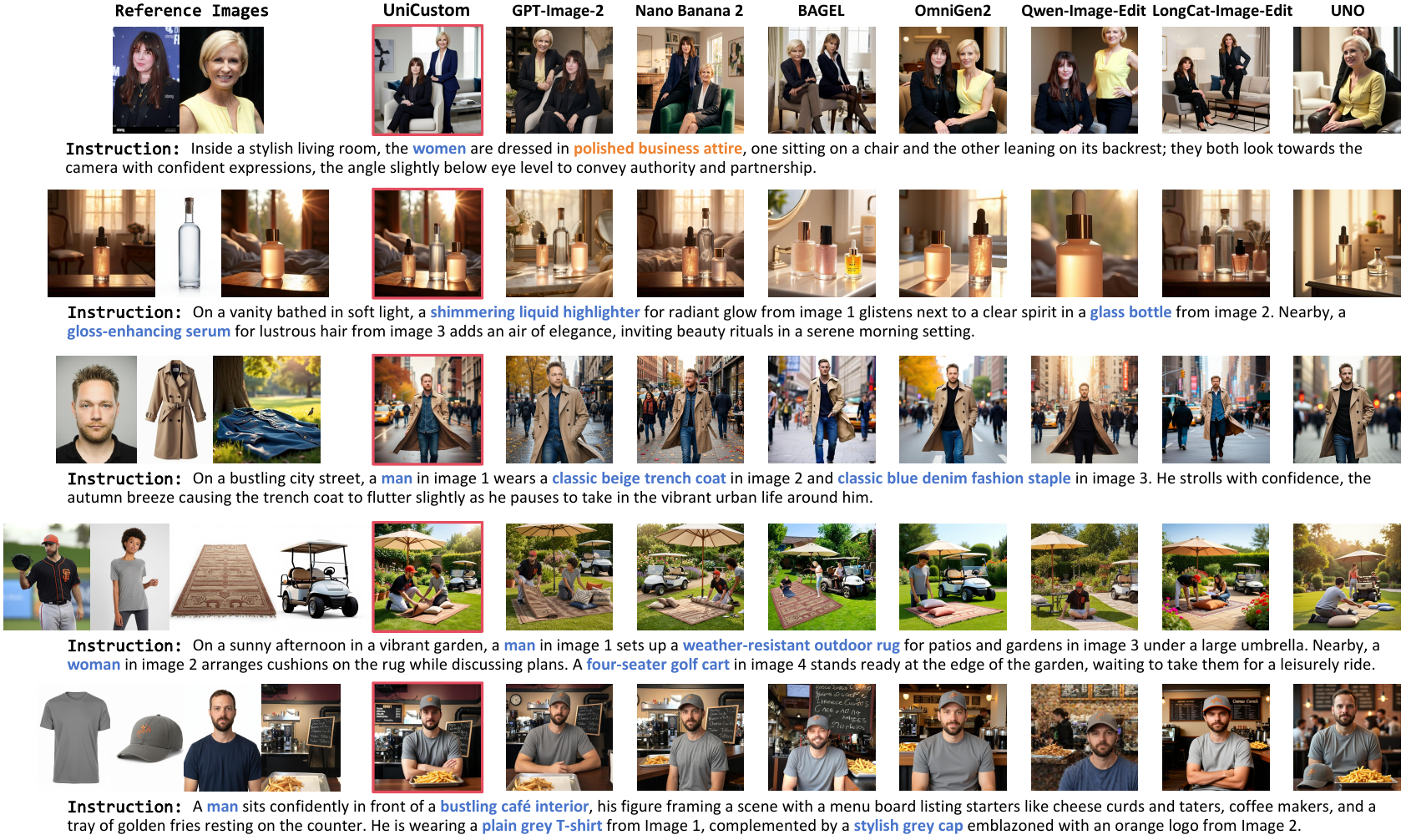}
    \caption{\textbf{Qualitative comparison on MICo-Bench~\cite{wei2025mico}.}}
    \label{fig:case_mico}
\end{figure}

\paragraph{Qualitative evaluation.}
We provide qualitative comparisons with competitive baselines on OmniContext and MICo-Bench in Figure~\ref{fig:case_omnicontext} and Figure~\ref{fig:case_mico}, covering challenging multi-reference scenarios involving humans, objects, clothing, scenes, and spatial relationships. Compared with existing baselines, UniCustom better preserves the visual identities and fine-grained attributes from the reference images while following complex textual instructions to model diverse relationships among multiple references, including subject interactions, spatial arrangements, role assignments, and object-attribute correspondences. For example, UniCustom can faithfully generate interactions such as hugging or facing each other, spatial layouts such as one subject sitting while another leans on the chair backrest, and multi-reference scenes involving people and objects with coherent spatial arrangements. These results demonstrate the superiority of UniCustom in complex multi-reference generation, where it better avoids subject missing and attribute confusion while achieving stronger instruction following than competing methods. More cases can be found in Figure~\ref{fig:case_omnicontext_appendix} and Figure~\ref{fig:case_micobench_appendix} in the Appendix.

\begin{wrapfigure}{r}{0.5\textwidth}
\vspace{-1em}
\centering
\includegraphics[width=\linewidth]{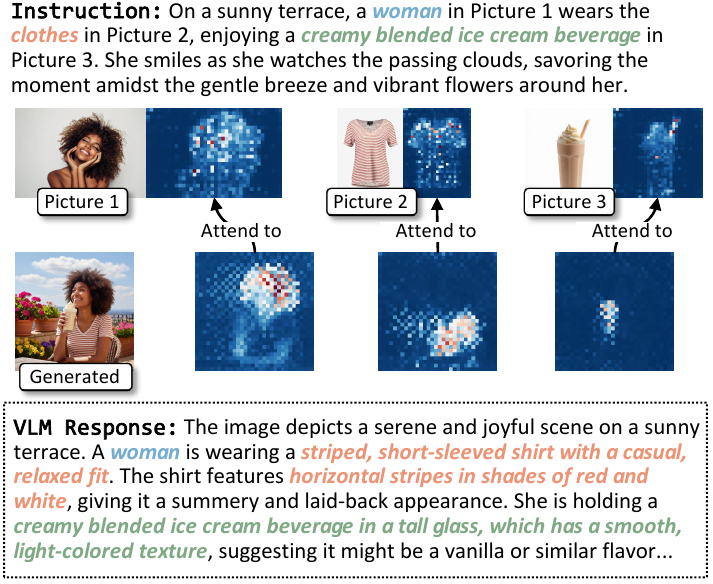}
\vspace{-2em}
\caption{\textbf{Attention visualization of UniCustom.}}
\vspace{-1em}
\label{fig:attn}
\end{wrapfigure}

\subsection{Reference Grounding and Binding Analysis}

To further analyze how UniCustom leverages multiple visual references, we visualize the internal attention maps of the DiT in multi-reference image generation. As shown in Figure~\ref{fig:attn}, given references containing a woman, a striped shirt, and a creamy blended beverage, UniCustom accurately attends to the corresponding regions in different reference images when generating each target component, demonstrating the effectiveness of UniCustom in semantic grounding and visual binding across multiple reference images.
We also present the corresponding VLM response, which correctly identifies the woman, the striped short-sleeved shirt, and the creamy blended beverage, indicating that the fused VAE and ViT features are well understood by the VLM and support accurate instruction parsing. These results show that UniCustom can reliably associate textual expressions with their visual references and generate coherent outputs following complex multi-image instructions.

\subsection{Ablation Study}

We conduct ablations in the pretraining stage to validate two key design choices: slot-wise binding regularization and early fusion. We report single-image reconstruction Signal-to-Noise Ratio (PSNR)~\cite{hore2010image}, and multi-image reconstruction, localization, and tiling accuracies, where the accuracies evaluate only image selection and placement rather than pixel-level reconstruction fidelity. These metrics measure whether the model can preserve visual details, bind them to parseable hidden states, and correctly use multiple input images, which are essential for multi-reference image generation; failures at pretraining will directly limit downstream generation quality.

\begin{figure}[t]
    \centering
    \setlength{\abovecaptionskip}{0.cm}
    \setlength{\belowcaptionskip}{-0.2cm}
    \includegraphics[width=0.85\linewidth]{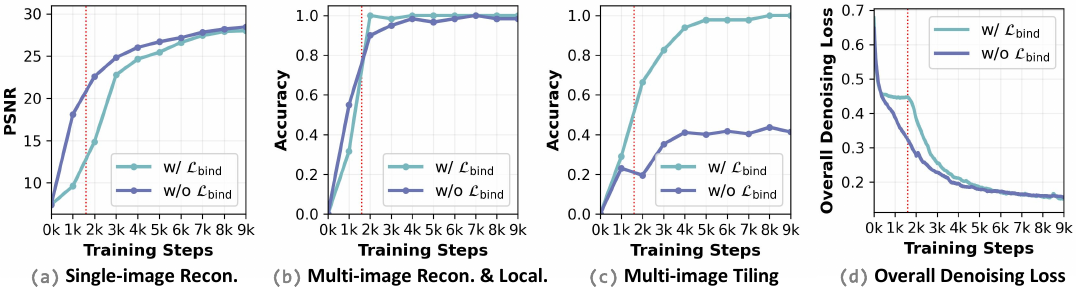}
    \caption{\textbf{Effect of slot-wise binding regularization}, where ``Recon.'' and ``Local.'' denote ``Reconstruction'' and ``Localization'', respectively.}
    \label{fig:ablation_binding}
\end{figure}

\paragraph{Effect of slot-wise binding regularization.}
\label{sec:regularization}
Figure~\ref{fig:ablation_binding} reveals distinct learning dynamics with and without slot-wise binding regularization during pretraining. 
Without $\mathcal{L}_{\mathrm{bind}}$, the model quickly learns to pass through VAE-level visual details, leading to faster early improvement in single-image reconstruction as shown in Figure~\ref{fig:ablation_binding}(a). 
However, this early visual-detail transmission does not translate into effective multi-image indexing, as reflected by slow learning in multi-image selection and consistently poor tiling accuracy in Figure~\ref{fig:ablation_binding}(b) and (c).
We attribute this behavior to the entangled multi-image hidden states produced by decoder-only VLMs. In models such as Qwen2.5-VL, the causal attention mechanism allows each token position to aggregate information from preceding tokens. When multiple images are serialized into a single sequence, visual details from different images can become mixed in the resulting hidden states, rather than remaining image-wise separable. Since these hidden states serve as the conditioning signals for the DiT, such entanglement makes it difficult to extract the relevant visual details needed for generation.

Slot-wise binding regularization alleviates this issue by encouraging localized image-slot binding, thereby producing more structured and parseable hidden states for resolving multi-image inputs.
With $\mathcal{L}_{\mathrm{bind}}$, each image slot is explicitly regularized to bind to the visual details of its corresponding image. Therefore, the model first learns a structured slot-image correspondence before substantial reconstruction improvement emerges. This explains why the generation loss changes only marginally in the early stage, especially before 2K steps, as shown in Figure~\ref{fig:ablation_binding}(d). Once this indexing structure is established, the DiT can parse the VLM hidden states more effectively, leading to a sharp loss decrease and clear gains in multi-image reconstruction, localization, and tiling accuracy.

\begin{figure}[t]
    \centering
    \setlength{\abovecaptionskip}{0.cm}
    \setlength{\belowcaptionskip}{-0.2cm}
    \includegraphics[width=\linewidth]{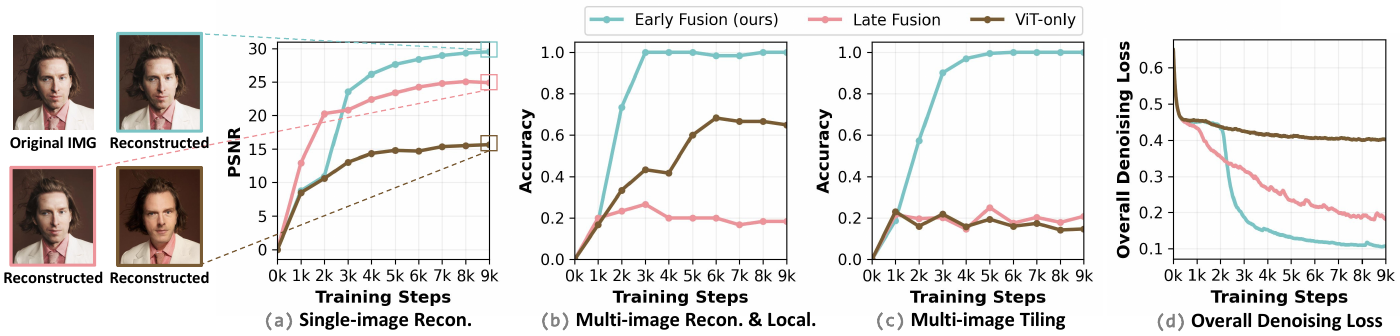}
    \caption{\textbf{Effect of different fusion strategies}, where ``Recon.'' and ``Local.'' denote ``Reconstruction'' and ``Localization'', respectively.}
    \label{fig:ablation_fusion}
\end{figure}

\paragraph{Effect of different fusion strategies.} We compare our early fusion strategy with two variants during pretraining: late fusion, which injects VAE features into the corresponding image slots after VLM encoding, and ViT-only, which excludes VAE features entirely. As shown in Figure 8(a), late fusion substantially improves single-image reconstruction over ViT-only and converges faster in the early training stage, indicating that directly injected VAE features can effectively provide fine-grained appearance information to the DiT. However, this improvement in reconstruction fidelity does not translate into reliable multi-image selection. Surprisingly, late fusion performs even worse than ViT-only on multi-image reconstruction and localization, as shown in Figure~\ref{fig:ablation_fusion}(b), while both variants exhibit consistently low tiling accuracy in Figure~\ref{fig:ablation_fusion}(c). Although late fusion explicitly assigns appearance-rich VAE features to their corresponding image slots, these features do not participate in the VLM's contextualization of visual references and textual instructions. Consequently, the encoded VLM hidden states are formed without access to fine-grained appearance information, leaving the DiT to coordinate separately encoded semantic and appearance cues. This separation may hinder effective reference grounding and visual information utilization. In contrast, early fusion achieves both the highest reconstruction quality and nearly perfect multi-image reconstruction, localization, and tiling accuracy. Together, these results suggest that effective multi-reference generation requires not only access to fine-grained appearance details, but also their integration into the VLM's contextualized representations, highlighting the importance of unified visual conditioning for bridging the grounding–binding gap.

%% file: tables/omnicontext.tex
\begin{table}[t]
\setlength{\abovecaptionskip}{-0.1cm}
\setlength{\belowcaptionskip}{0.1cm}
\caption{\textbf{Quantitative comparison on OmniContext}~\cite{wu2025omnigen2}, where ``Char.'' and ``Obj.'' denote ``Character'' and ``Object'', respectively. The best results in each group are highlighted in bold.}
\label{tab:omnicontext}
\renewcommand{\arraystretch}{1.2}
\setlength\tabcolsep{4pt}  
\resizebox{\textwidth}{!}{
\begin{tabular}{lccccccccc}
\toprule
                                              & \multicolumn{2}{c}{\textbf{SINGLE}}                                         & \multicolumn{3}{c}{\textbf{MULTIPLE}}                                                                              & \multicolumn{3}{c}{\textbf{SCENE}}                                                                                 &                                      \\ \cline{2-9}
\multirow{-2}{*}{\textbf{\# OmniContext}}     & \textbf{Character}                   & \textbf{Object}                      & \textbf{Character}                   & \textbf{Object}                      & \textbf{Char.+Obj.}            & \textbf{Character}                   & \textbf{Object}                      & \textbf{Char.+Obj.}            & \multirow{-2}{*}{\textbf{AVERAGE}}   \\ \midrule
\multicolumn{10}{c}{\cellcolor[HTML]{E3EEF8}\textbf{Closed-source Models}}                                                                                                                                                                                                                                                                                                                                   \\ \midrule
GPT-Image-2~\cite{gpt-image-2}                                   & {\color[HTML]{2C2F33} \textbf{9.24}} & {\color[HTML]{2C2F33} \textbf{9.65}} & {\color[HTML]{2C2F33} \textbf{9.13}} & {\color[HTML]{2C2F33} \textbf{9.50}} & {\color[HTML]{2C2F33} \textbf{8.94}} & {\color[HTML]{2C2F33} \textbf{9.27}} & {\color[HTML]{2C2F33} \textbf{9.25}} & {\color[HTML]{2C2F33} \textbf{8.95}} & {\color[HTML]{2C2F33} \textbf{9.24}} \\
Nano Banana 2~\cite{nanobanan}                                 & 8.98                                 & 9.44                                 & 8.85                                 & 9.26                                 & 8.81                                 & 8.80                                 & 8.59                                 & 8.67                                 & 8.92                                 \\ \midrule
\multicolumn{10}{c}{\cellcolor[HTML]{E4E4F8}\textbf{Open-source Models}}                                                                                                                                                                                                                                                                                                                                     \\ \midrule
{\color[HTML]{1F2329} FLUX-Kontext {[}dev{]}}~\cite{labs2025flux} & {\color[HTML]{1F2329} 7.09}          & {\color[HTML]{1F2329} 7.35}          & {\color[HTML]{1F2329} 2.49}          & {\color[HTML]{1F2329} 5.46}          & {\color[HTML]{1F2329} 4.46}          & {\color[HTML]{1F2329} 2.86}          & {\color[HTML]{1F2329} 3.89}          & {\color[HTML]{1F2329} 3.40}          & {\color[HTML]{1F2329} 4.62}          \\
{\color[HTML]{1F2329} UNO~\cite{wu2025less}}                    & {\color[HTML]{1F2329} 7.51}          & {\color[HTML]{1F2329} 7.99}          & {\color[HTML]{1F2329} 4.56}          & {\color[HTML]{1F2329} 7.41}          & {\color[HTML]{1F2329} 6.53}          & {\color[HTML]{1F2329} 4.00}          & {\color[HTML]{1F2329} 5.76}          & {\color[HTML]{1F2329} 5.77}          & {\color[HTML]{1F2329} 6.19}          \\
{\color[HTML]{1F2329} USO~\cite{wu2025uso}}                    & {\color[HTML]{1F2329} 7.71}          & {\color[HTML]{1F2329} 7.68}          & {\color[HTML]{1F2329} 2.91}          & {\color[HTML]{1F2329} 7.27}          & {\color[HTML]{1F2329} 5.61}          & {\color[HTML]{1F2329} 3.88}          & {\color[HTML]{1F2329} 6.49}          & {\color[HTML]{1F2329} 6.04}          & {\color[HTML]{1F2329} 5.95}          \\
{\color[HTML]{1F2329} BAGEL~\cite{deng2025emerging}}                  & {\color[HTML]{1F2329} 7.12}          & {\color[HTML]{1F2329} 7.70}          & {\color[HTML]{1F2329} 5.86}          & {\color[HTML]{1F2329} 7.07}          & {\color[HTML]{1F2329} 6.79}          & {\color[HTML]{1F2329} 4.98}          & {\color[HTML]{1F2329} 6.05}          & {\color[HTML]{1F2329} 6.03}          & {\color[HTML]{1F2329} 6.45}          \\
{\color[HTML]{1F2329} OmniGen2~\cite{wu2025omnigen2}}               & {\color[HTML]{1F2329} 8.04}          & {\color[HTML]{1F2329} 7.87}          & {\color[HTML]{1F2329} 7.22}          & {\color[HTML]{1F2329} 7.76}          & {\color[HTML]{1F2329} 7.51}          & {\color[HTML]{1F2329} 7.52}          & {\color[HTML]{1F2329} 7.30}          & {\color[HTML]{1F2329} 7.57}          & {\color[HTML]{1F2329} 7.60}          \\
Qwen-Image-Edit-2509~\cite{wu2025qwen}                          & 8.11                                 & \textbf{9.01}                        & \textbf{7.85}                        & \textbf{8.53}                        & 7.60                                 & 5.88                                 & 7.38                                 & 7.12                                 & 7.69                                 \\
LongCat-Image-Edit~\cite{team2025longcat}                            & \textbf{8.24}                        & 8.63                                 & 6.69                                 & 8.07                                 & 6.99                                 & 5.80                                 & 6.73                                 & 6.98                                 & 7.26                                 \\ \midrule
UniCustom (ours)                              & 8.06                                 & 7.51                                 & 7.78                                 & 7.99                                 & \textbf{7.86}                        & \textbf{7.86}                        & \textbf{7.72}                        & \textbf{7.92}                        & \textbf{7.84}                        \\ \bottomrule
\end{tabular}
}
\end{table}

%% file: tables/micobench.tex
\begin{wraptable}{r}{0.5\textwidth}
\vspace{-1em}
\setlength{\belowcaptionskip}{0.1cm}
\caption{\textbf{Quantitative comparison on MICo-Bench}~\cite{wei2025mico}, where ``HOI'' and ``De\&Re'' denote ``Human-object Interaction'' and ``Decomposition \& Recomposition'', respectively. The best results in each group are highlighted in bold.}
\label{tab:micobench}
\resizebox{\hsize}{!}{
\renewcommand{\arraystretch}{1.30}  
\setlength\tabcolsep{2pt}  
\begin{tabular}{lccccc}
\toprule
\textbf{\# MICo-Bench}                        & \textbf{Object} & \textbf{Person} & \textbf{HOI}   & \textbf{De\&Re} & \textbf{Overall} \\ \midrule
\multicolumn{6}{c}{\cellcolor[HTML]{E3EEF8}\textbf{Closed-source Models}}                                                               \\ \midrule
GPT-Image-2~\cite{gpt-image-2}                                   & 66.93           & 60.56           & 59.02          & 62.69           & 61.78            \\
Nano Banana 2~\cite{nanobanan}                                 & \textbf{68.46}  & \textbf{65.88}  & \textbf{64.88} & \textbf{70.28}  & \textbf{67.42}   \\ \midrule
\multicolumn{6}{c}{\cellcolor[HTML]{E4E4F8}\textbf{Open-source Models}}                                                                 \\ \midrule
{\color[HTML]{1F2329} FLUX-Kontext {[}dev{]}~\cite{labs2025flux}} & 21.40           & 14.33           & 12.67          & 7.24            & 12.51            \\
{\color[HTML]{1F2329} UNO~\cite{wu2025less}}                    & 42.20           & 15.15           & 27.99          & 41.93           & 32.43            \\
{\color[HTML]{1F2329} USO~\cite{wu2025uso}}                    & 38.18           & 15.57           & 20.25          & 35.91           & 27.37            \\
BAGEL~\cite{deng2025emerging}                                         & 39.23           & 16.77           & 20.99          & 46.75           & 31.62            \\
Qwen-Image-Edit-2509~\cite{wu2025qwen}                          & 37.33           & \textbf{28.01}  & 22.06          & 10.53           & 21.67            \\
LongCat-Image-Edit~\cite{team2025longcat}                            & 40.55           & 19.40           & 24.47          & 14.86           & 22.78            \\ \midrule
UniCustom (ours)                              & \textbf{54.30}  & 18.12           & \textbf{40.51} & \textbf{50.29}  & \textbf{41.71}   \\ \bottomrule
\end{tabular}
}
\end{wraptable}

%% file: 4_related_work.tex
\section{Related Work}

\paragraph{Multi-reference Image Generation.}
 
To enable multi-reference image generation, early methods mainly rely on per-subject optimization~\cite{wang2025msdiffusion} or adapter-based feature injection~\cite{ye2023ip,li2023blipdiffusion}. More recently, in-context conditioning has emerged as a more flexible paradigm. OminiControl~\cite{tan2025ominicontrol} shows that DiT can inherently encode visual references, motivating methods such as UNO~\cite{wu2025less}.
Subsequent works further improve this paradigm through attention constraints, as in MOSAIC~\cite{she2026mosaic} and DreamO~\cite{mou2025dreamo}; modulation-based control, as in TokenVerse~\cite{garibi2025tokenverse} and XVerse~\cite{chen2026xverse}; and reinforcement learning, as in UMO~\cite{cheng2025umo} and PSR~\cite{wang2025psr}. Building on this trend, VLM-enhanced methods such as OmniGen2~\cite{wu2025omnigen2}, Qwen-Image-Edit~\cite{wu2025qwen}, and Canvas-to-Image~\cite{dalva2025canvas} leverage VLMs to improve instruction understanding and reference-aware generation. UniCustom further advances VLM-enhanced conditioning by unifying semantic ViT features and appearance-rich VAE features within the VLM conditioning stream, rather than processing them through decoupled pathways, yielding unified conditioning signals that jointly capture subject-level cues and reference-specific visual details for faithful generation.

\paragraph{VLM-enhanced Diffusion Models for Image-to-Image Generation.}
VLM-enhanced diffusion models have recently advanced image-to-image generation by incorporating multimodal understanding into diffusion-based synthesis. Some methods use VLMs to provide auxiliary guidance for diffusion models, such as input image encoding in BLIP-Diffusion~\cite{li2023blipdiffusion}, multimodal prompt understanding in Kosmos-G~\cite{pan2024kosmosg}, and instruction refinement in MGIE~\cite{fu2024guiding}. Others move toward unified multimodal generation and editing frameworks, including Step1X-Edit~\cite{liu2025step1x}, OmniGen2~\cite{wu2025omnigen2}, Qwen-Image-Edit~\cite{wu2025qwen}, and LongCat-Image-Edit~\cite{team2025longcat}, where instruction understanding and image-to-image synthesis are integrated in a single system. While these methods differ in architecture and scope, they commonly rely on separate pathways for semantic visual features and appearance-rich generative features. This design leaves the correspondence between textual instructions and low-level visual details to be inferred implicitly during generation. UniCustom differs by explicitly aligning semantic and generative visual information before VLM encoding, yielding representations that are both semantically grounded and appearance-aware.

%% file: 5_conclusion.tex
\section{Conclusion}

We presented \textbf{\textit{UniCustom}}, a unified visual conditioning framework for multi-reference image generation that addresses the grounding--binding gap in existing VLM-enhanced diffusion models. By fusing semantic ViT features and appearance-rich VAE features before VLM encoding, UniCustom enables the resulting hidden states to jointly encode instruction-level subject cues and reference-specific visual details. Combined with a two-stage training strategy and slot-wise binding regularization, this design improves subject consistency, instruction following, and compositional fidelity in complex multi-reference scenarios. Experiments on OmniContext and MICo-Bench demonstrate that UniCustom achieves strong performance among open-source methods. More broadly, our results show that appearance-rich VAE information can be effectively propagated through a frozen VLM and leveraged by the diffusion backbone for faithful generation. This suggests a promising direction for designing unified visual representations that connect multimodal understanding and generation.

%% file: acknowledgement.tex
\section*{Acknowledgements and Disclosures}

This work was supported by the National Natural Science Foundation of China (U25A20340, U25B2071). This work was conducted during the first author's internship at KlingAI, which provided computational resources for the research. The authors declare no other competing interests.

%% file: appendix.tex
\section{Implementation Details}
\label{sec:implementation_appendix}
During pretraining of UniCustom, we jointly optimize the fusion layer and the DiT with a learning rate of $5\times10^{-5}$ for 18K steps. The training batches are uniformly sampled from four task groups: multi-image reconstruction, localization, tiling, and understanding. For supervised finetuning, we freeze the fusion layer and only update the DiT. We train for another 18K steps with a learning rate of $1\times10^{-5}$. The finetuning mixture consists of 10\% pretraining tasks, 5\% text-to-image generation, 10\% image editing, 25\% single-reference generation, and 50\% multi-reference generation. This mixture preserves the reference grounding ability acquired during pretraining while adapting the model to open-ended reference-based image generation. The model is trained using 128 GPUs.

Note that OmniGen2~\cite{wu2025omnigen2} is omitted from the quantitative comparison on MICo-Bench in Table~\ref{tab:micobench} since it does not support generation conditioned on more than five reference images.

\section{Training Data}
\label{sec:data}
\paragraph{Multi-image reconstruction.} We first curate a large-scale image pool from open-source datasets, including HQ-50K~\cite{yang2023hq}, MultiID-2M~\cite{xu2026withanyone}, HumanArt~\cite{ju2023human}, FFHQ~\cite{karras2019style}, and Unsplash~\cite{unsplash_datasets_github}. 
Based on this corpus, we construct multi-image reconstruction data for pretraining. Each sample consists of multiple candidate images paired with an instruction that specifies the target image to be reconstructed (\eg ``\textit{Reconstruct Picture $i$ with the highest possible fidelity\dots}''). To avoid resolution-based shortcuts, we partition images into resolution buckets and sample candidates from the same bucket, which encourages the model to ground the target image from the specified image identifier, rather than exploiting superficial resolution cues.

\paragraph{Multi-image localization.}
We construct multi-image localization data from COCO2017~\cite{lin2014microsoft}. 
Starting from single-image bounding-box and segmentation samples, 
we extend them into multi-image localization tasks following the same protocol as multi-image reconstruction. Each sample contains multiple candidate images with the same resolution, and the instruction specifies the target image to be localized (\eg ``\textit{Please bound the woman in Picture $i$}'', ``\textit{Segment the boats in Picture $i$.}''), which strengthens semantic binding while preserving localization ability. By requiring localization under an explicitly specified reference index, this task promotes semantic grounding between the instruction and the target image, while reinforcing visual binding between the selected reference and its localized content.

\paragraph{Multi-image tiling.}
Based on the image pool used for multi-image reconstruction, we further construct multi-image tiling data. Specifically, we sample multiple images from the same resolution bucket and arrange them into grid-based collages with layouts of $2\times2$, $2\times3$, $3\times2$, and $3\times3$. The instruction specifies both the canvas layout and the ordering of input images (\eg ``\textit{Construct a collage on a $2\times3$ canvas. Fill the grid using images in this order: Picture 5,\dots, Picture 1. Place them sequentially from left to right and then top to bottom\dots}''). This task requires the model to follow explicit ordering constraints across multiple images, thereby strengthening semantic grounding to the image identifier and visual binding between each image and its assigned spatial position.

\paragraph{Multi-image understanding (auxiliary task).}
We construct multi-image understanding data from the captioning and general VQA subsets of Honey-Data-15M~\cite{zhang2025bee}. Following the same multi-image augmentation protocol, we convert single-image samples into multi-image tasks by adding candidate images and rewriting the instruction to query a specified reference (\eg ``\textit{Answer the question based on Picture $i$\dots}'') This promotes semantic grounding to the designated image while retaining image-level understanding ability in multi-reference scenarios.

\paragraph{Single- and multi-reference generation.}
We collect reference-based generation data from Echo-4o-Image~\cite{ye2025echo}, Nano-Consistent-150K~\cite{ye2025echo}, MICo-150K~\cite{wei2025mico}, and an internal dataset. After filtering and reformatting, 
these data further adapt the model from reconstruction- and understanding-oriented pretraining to reference-based generation, where the model must ground textual instructions to the specified references and preserve the corresponding visual identities during synthesis.

\paragraph{Text-to-image generation and image editing (auxiliary task).}
Following UniWorld-V1~\cite{lin2025uniworld}, we include a small amount of text-to-image data from BLIP-3o~\cite{chen2025blip3} and Open-Sora Plan~\cite{lin2024open}, with image editing data from Pico-Banana-400K~\cite{qian2025pico}, ImgEdit~\cite{ye2025imgedit}, and NHR-Edit~\cite{kuprashevich2026nohumansrequired}. These samples serve as auxiliary SFT data that broaden the instruction distribution beyond reference-based prompts, while the training objective remains focused on single- and multi-reference image generation.

\begin{figure*}[!t]
\centering
\begin{tcolorbox}[
  width=\textwidth,
  colback=white,
  colframe=gray!70,
  enhanced,
  sharp corners,
  boxrule=0.5pt,
  drop shadow,
  fonttitle=\bfseries\small,
  coltitle=black
]
\footnotesize
This survey evaluates images generated by AI for a multi-reference image generation task. 

\medskip
The survey contains 60 questions. In each question, you will see:\\
1. A text prompt describing the requested image;\\
2. One or more reference images showing the subjects or visual elements that should appear in the result;\\
3. Three AI-generated images labeled A, B, and C.\\
Please select the generated image that best follows the text prompt while preserving the identity and visual appearance of the subjects in the reference images.

\medskip
\textbf{\# Evaluation Criteria}\\
Please consider the following criteria:\\
\textbf{1. Prompt Following:} How well the generated image follows the requested content, actions, relationships, composition, and setting.\\
\textbf{2. Subject Consistency:} How accurately the subjects preserve the identity and visual appearance shown in the reference images.\\
\textbf{3. Image Quality:} Whether the image is coherent, natural-looking, detailed, and free of visible artifacts.\\
\textbf{4. Aesthetics:} The overall visual appeal of the image, including its composition, lighting, color, and style.

\medskip
Please give greater importance to Subject Consistency and Prompt Following when making your choice. Select one answer only for each question. There are no right or wrong answers — we are interested in your visual judgment. Please complete the survey using a desktop or laptop computer so that you can examine the images carefully.

\end{tcolorbox}

\vspace{-2mm}
\caption{\textbf{Instructions for human evaluation.}}
\vspace{-2mm}
\label{box:human_eval_instruction}
\end{figure*}

\begin{figure}[h]
    \centering
    \setlength{\abovecaptionskip}{0.2cm}
    \setlength{\belowcaptionskip}{-0.2cm}
    \includegraphics[width=0.6\linewidth]{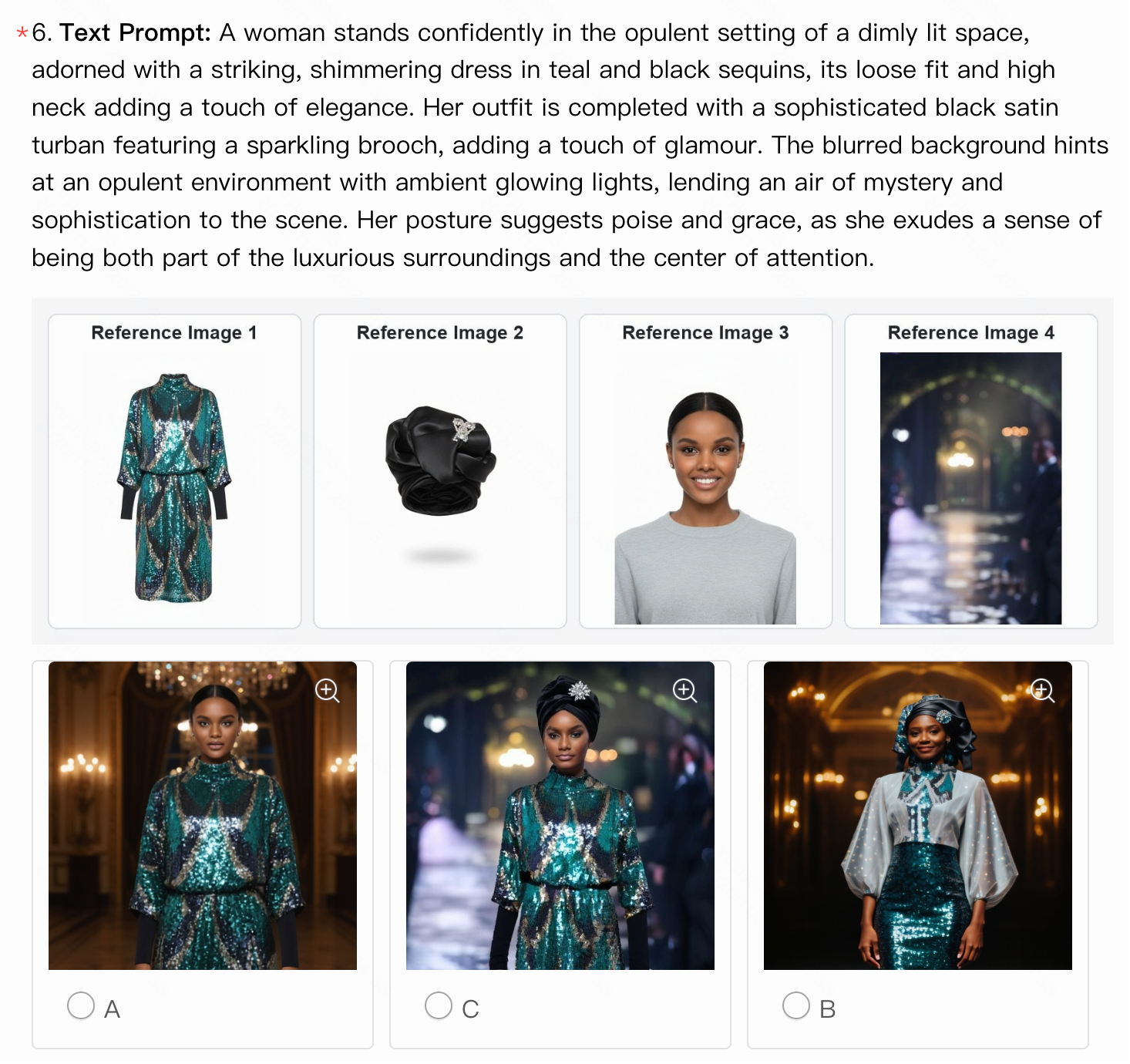}
    \caption{\textbf{Demonstration of human evaluation interface.}}
    \label{fig:human_eval_interface}
\end{figure}

\input{tables/human_eval}

\section{Human Evaluation}
\label{sec:human_eval}
We conducted a human evaluation on 60 cases, including 30 from OmniContext and 30 from MICo-Bench. We recruited 50 participants through CloudResearch~\footnote{https://www.cloudresearch.com/.} to evaluate all 60 cases, yielding 3,000 judgments in total. For each case, outputs from UniCustom, Qwen-Image-Edit, and LongCat-Image-Edit were presented in randomized order. Participants selected the best one by jointly considering prompt following, subject consistency, overall visual quality, and aesthetics. The full text of instructions and the evaluation interface are presented in Figure~\ref{box:human_eval_instruction} and Figure~\ref{fig:human_eval_interface}. The resulting preference rates with 95\% confidence intervals are presented in Table~\ref{tab:human_eval}. UniCustom received the highest overall preference, which is consistent with the quantitative results.

\section{Feature Propagation Analysis}
Although UniCustom achieves high reconstruction fidelity through unified visual conditioning, it remains unclear how fine-grained VAE information is preserved and propagated through the frozen VLM. To better understand this process, we analyze the representation alignment and information recoverability of image-slot hidden states across different VLM layers.

\paragraph{Representation alignment.} We first compute linear Centered Kernel Alignment (CKA) between intermediate VLM hidden states and the original ViT and VAE features on $200$ held-out images. As a control, we also evaluate ViT-only features processed by the same frozen VLM. As presented in Table~\ref{tab:cka}, the fused representations initially exhibit high alignment with ViT features, consistent with our identity-preserving initialization. Interestingly, their alignment with VAE features progressively increases from $0.541$ before VLM encoding to $0.938$ at the final layer, whereas ViT-only representations maintain consistently low alignment. This suggests that the learned fusion layer enables VAE information to be effectively propagated through the frozen VLM.

\input{tables/cka}

\paragraph{Representation recoverability.} Since CKA measures representational similarity rather than information recoverability, we further train layer-wise linear probes to predict VAE latents from intermediate hidden states, which are then decoded into images using the frozen VAE decoder. We report the reconstruction PSNR on $500$ held-out images. As shown in Table~\ref{tab:probe}, the fused representations achieve a PSNR of $29.00$ dB, substantially outperforming ViT-only representations and approaching the VAE reconstruction upper bound ($33.26$ dB). These results indicate that fine-grained appearance information remains linearly recoverable after VLM encoding, supporting the effectiveness of unified visual conditioning in preserving appearance-aware information for DiT generation.

\input{tables/probe}

\section{Qualitative Results}
\begin{figure}[h]
    \centering
    \setlength{\abovecaptionskip}{0.2cm}
    \setlength{\belowcaptionskip}{-0.2cm}
    \includegraphics[width=0.6\linewidth]{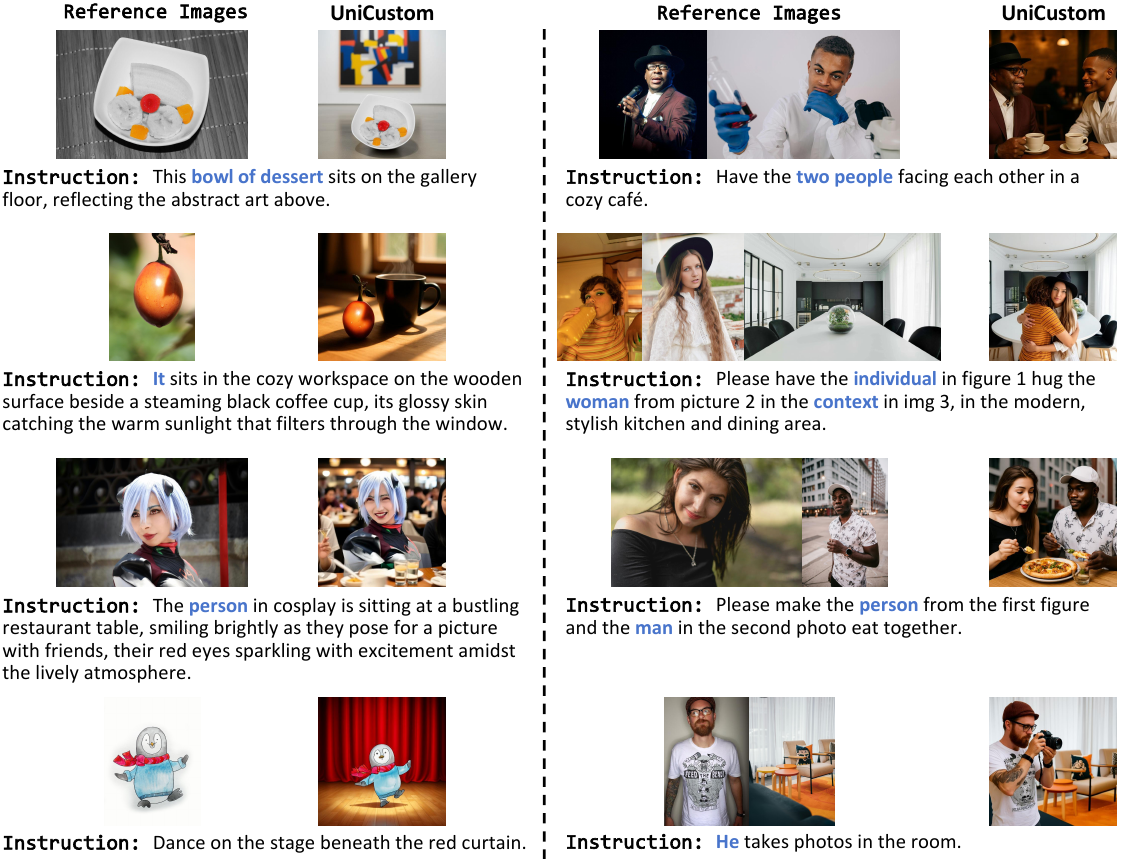}
    \caption{\textbf{More generated examples of UniCustom on OmniContext~\cite{wu2025omnigen2}.}}
    \label{fig:case_omnicontext_appendix}
\end{figure}

\begin{figure}[h]
    \centering
    \setlength{\abovecaptionskip}{0.2cm}
    \setlength{\belowcaptionskip}{-0.2cm}
    \includegraphics[width=\linewidth]{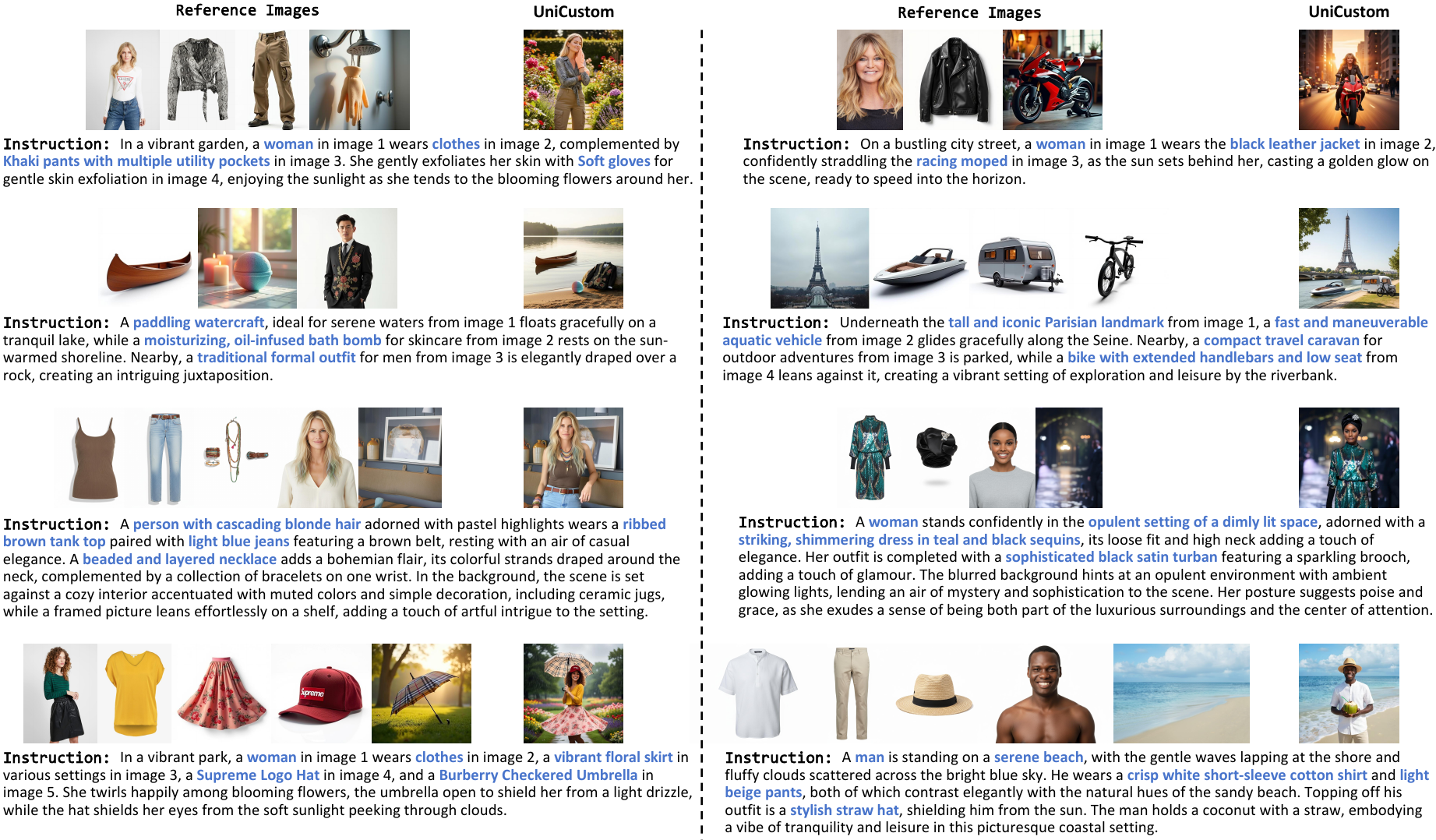}
    \caption{\textbf{More generated examples of UniCustom on MICo-Bench~\cite{wei2025mico}.}}
    \label{fig:case_micobench_appendix}
\end{figure}

\paragraph{Multi-reference image generation.}
We provide more generated examples of UniCustom on OmniContext~\cite{wu2025omnigen2} and MICo-Bench~\cite{wei2025mico} in Figure~\ref{fig:case_omnicontext_appendix} and Figure~\ref{fig:case_micobench_appendix}. These examples further validate the effectiveness of UniCustom in multi-reference image generation, showing that its unified visual conditioning helps preserve fine-grained reference appearances while maintaining accurate semantic grounding and coherent subject composition.

\begin{figure}[h]
    \centering
    \setlength{\abovecaptionskip}{0.2cm}
    \setlength{\belowcaptionskip}{-0.2cm}
    \includegraphics[width=\linewidth]{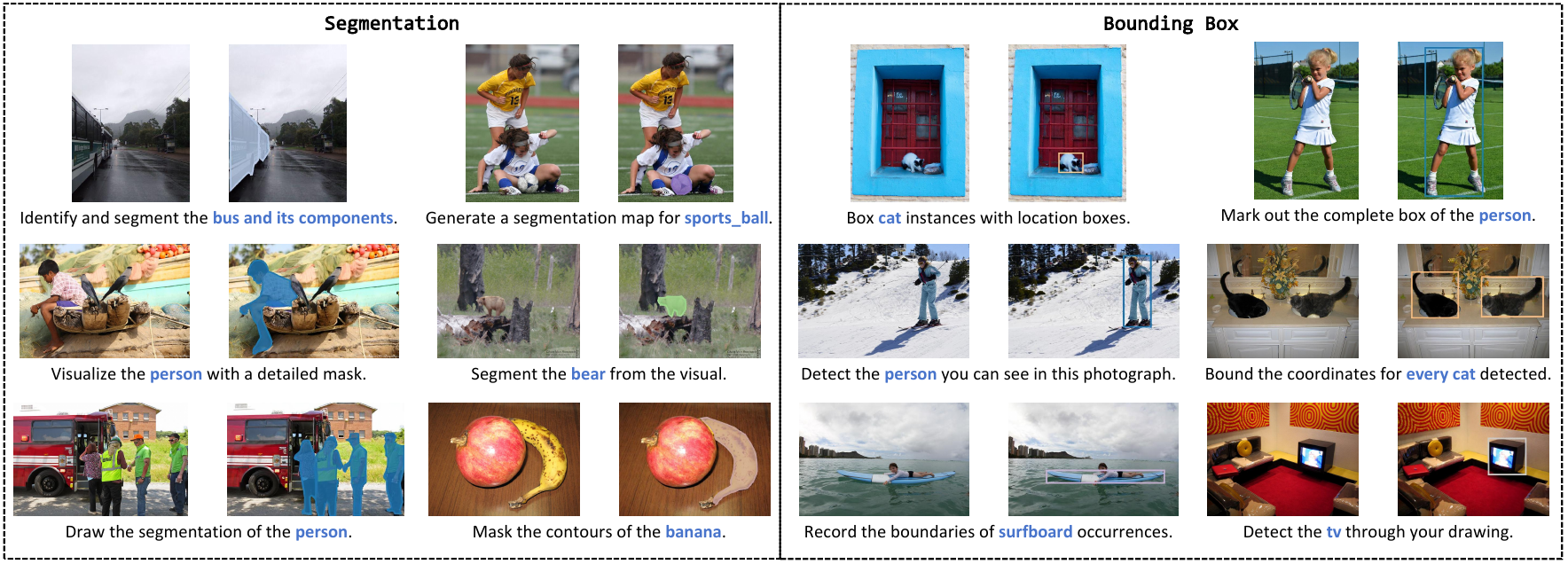}
    \caption{\textbf{Generated examples of UniCustom on image localization.}}
    \label{fig:case_localization}
\end{figure}

\paragraph{Image localization.}
As illustrated in Figure~\ref{fig:ablation_binding}, the single-image reconstruction PSNR demonstrates that low-level details encoded in the VAE features can be effectively transmitted through the VLM hidden states. In addition, the nearly perfect accuracy on multi-image reconstruction, localization, and tiling suggests that UniCustom establishes reliable semantic grounding and visual binding across multiple references. Specifically, the DiT can effectively parse the hidden states, follow the given instruction, and select or recombine the corresponding reference images as required. In Figure~\ref{fig:case_localization}, we further present qualitative results for image localization. These examples show that UniCustom can accurately follow instructions to localize the specified subject, which is a crucial capability for downstream multi-image reference tasks.

\begin{figure}[h]
    \centering
    \setlength{\belowcaptionskip}{-0.2cm}
    \includegraphics[width=0.75\linewidth]{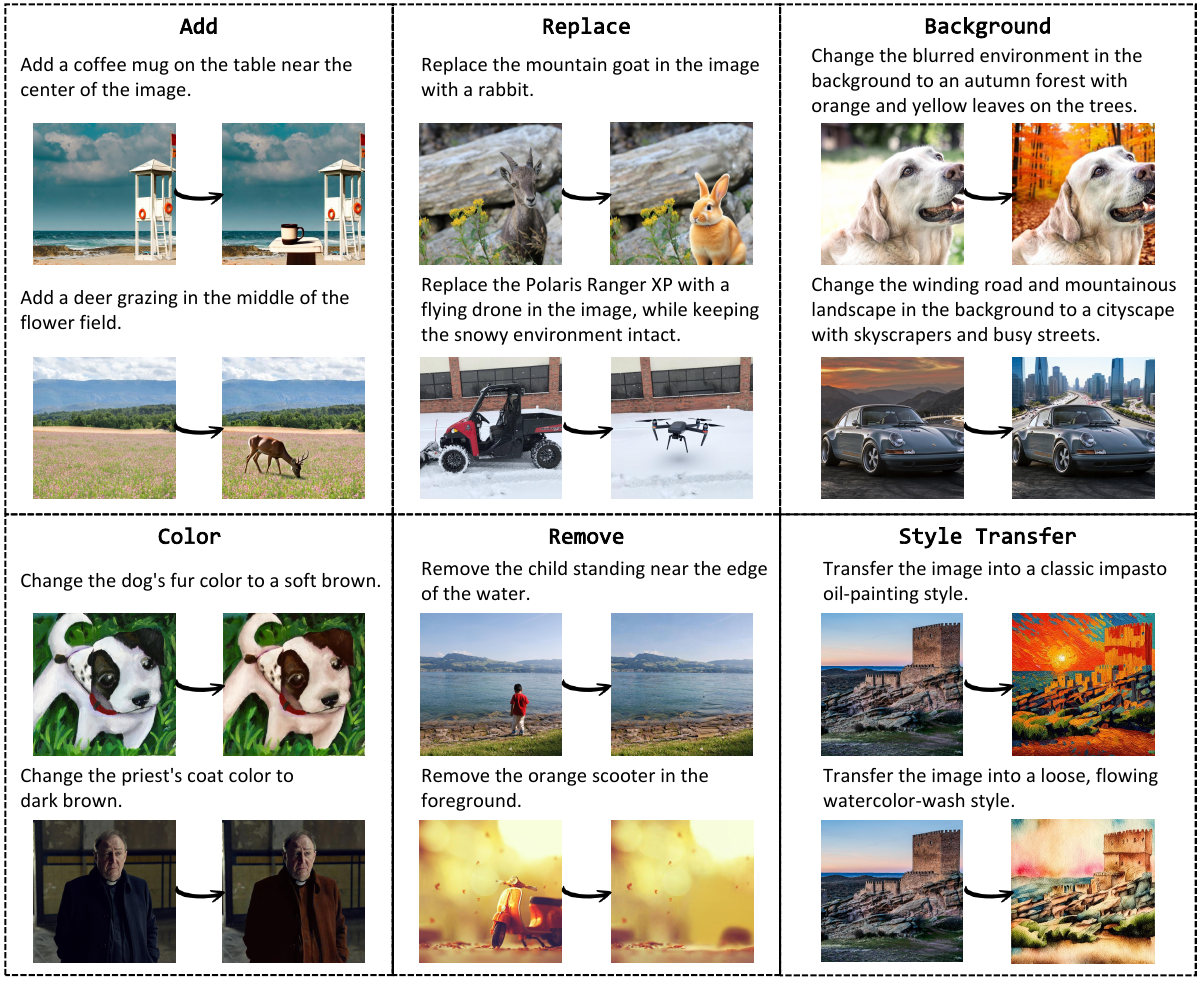}
    \caption{\textbf{Generated examples of UniCustom on image editing.}}
    \label{fig:case_edit}
\end{figure}

\begin{figure}[h]
    \centering
    \setlength{\abovecaptionskip}{0.2cm}
    \setlength{\belowcaptionskip}{-0.2cm}
    \includegraphics[width=\linewidth]{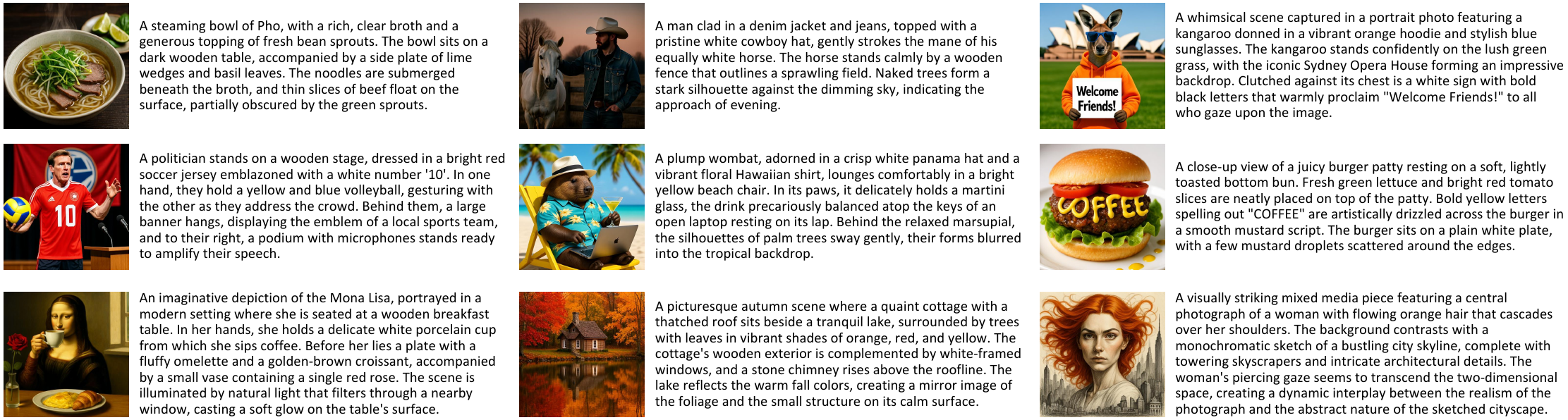}
    \caption{\textbf{Generated examples of UniCustom on text-to-image generation.}}
    \label{fig:case_t2i}
\end{figure}

\paragraph{Image editing and text-to-image generation.}
We further present qualitative results of UniCustom on image editing and text-to-image generation in Figures~\ref{fig:case_edit} and~\ref{fig:case_t2i}, respectively. Although UniCustom incorporates only a small amount of image editing and text-to-image data as auxiliary tasks to improve instruction adherence, it demonstrates strong generalization across both settings. For image editing, UniCustom performs well on a diverse range of tasks, including object addition, object replacement, background modification, color editing, object removal, and style transfer. For text-to-image generation, UniCustom also produces visually plausible and semantically aligned results, indicating that its learned capabilities extend beyond multi-reference generation.

\section{Limitations and Future Work}
\label{sec:limitation}

UniCustom achieves strong multi-reference image generation performance among open-source models, demonstrating the promise of unified visual conditioning for reference-based generation. Nevertheless, a gap remains compared with closed-source systems, particularly in highly realistic human identity preservation, complex object interactions, and fine-grained scene consistency. We regard this gap not as an inherent limitation of the framework, but as evidence of several important directions for future improvement.

First, UniCustom currently employs a lightweight fusion module that integrates spatially aligned ViT and VAE features through a simple linear layer. This design offers an initial exploration of combining understanding-oriented and generation-oriented visual representations within a unified conditioning pathway. Our results indicate that feature fusion before VLM encoding is effective; however, the current fusion mechanism remains relatively simple. Future work may investigate adaptive and hierarchical fusion strategies to better balance semantic grounding, appearance preservation, and spatial consistency.

Second, UniCustom is mainly trained and evaluated at a resolution no larger than $512 \times 512$. While this setting is sufficient for validating the core design, it may limit the preservation of small text, logos, subtle facial details, and intricate textures. Scaling UniCustom to higher-resolution generation is therefore an important next step, especially for applications that require realistic identity preservation and fine-grained visual fidelity.

Third, UniCustom is primarily designed for multi-reference generation rather than as a fully general-purpose multimodal generation system. Although our two-stage training pipeline incorporates image understanding, image editing, and text-to-image data, these tasks mainly serve the central objective of reference-based generation. Specifically, image understanding data in pretraining helps retain the VLM's instruction-following and semantic grounding abilities under the unified visual representation, while editing and text-to-image data in supervised finetuning improve instruction adherence. Consequently, UniCustom exhibits certain generalization ability on auxiliary tasks, as shown by its multi-reference instruction understanding in Figure~\ref{fig:attn}, image editing in Figure~\ref{fig:case_edit}, and text-to-image generation in Figure~\ref{fig:case_t2i}. However, its performance on these tasks remains moderate and is not yet comparable to models specifically optimized for them.

These limitations arise from multiple aspects of the current training recipe. For understanding-oriented tasks, the pretraining stage includes only a small proportion of image understanding data, mainly captioning and general VQA. Although such data helps preserve the instruction-following and multimodal understanding abilities, it is insufficient for more complex multimodal reasoning tasks, such as mathematics and science. For generation-oriented auxiliary tasks, the editing and text-to-image data used during supervised finetuning are limited in both scale and diversity compared with those used by task-specialized models. As a result, UniCustom can generalize to image editing and text-to-image generation to some extent, but these capabilities remain secondary to its main reference-generation objective. In addition, the VLM is kept frozen throughout training, leaving the lightweight fusion layer as the primary trainable component for adapting the unified visual representation. This design improves training efficiency and helps preserve the base VLM's capabilities, but it also constrains the model's adaptability to broader understanding and generation tasks.

Future work will explore higher-resolution generation, more expressive visual fusion modules, parameter-efficient or selective VLM adaptation, and improved training mixtures with more diverse and higher-quality data for understanding, editing, text-to-image generation, and reference-based generation. Overall, UniCustom represents an initial yet meaningful step toward unified visual conditioning for multimodal generation. Beyond its strong open-source multi-reference generation performance, our results suggest that fusing ViT- and VAE-derived visual representations before VLM encoding is a promising way to bridge understanding-oriented and generation-oriented visual signals. We hope this insight can support the development of a more general multimodal understanding and generation system.

\section{Ethical Statement}
\label{sec:ethics}
Our work focuses on algorithmic improvements for multi-reference image generation, aiming to improve subject consistency and compositional fidelity when synthesizing images from complex textual instructions and multiple reference images. The training data are mainly constructed from open-source datasets, with evaluation conducted on public benchmarks such as OmniContext~\cite{wu2025omnigen2} and MICo-Bench~\cite{wei2025mico}. No data are collected from human subjects. Although we do not anticipate direct or immediate negative societal impacts from the algorithmic contributions themselves, we recognize that improvements in image generation quality and controllability can amplify existing risks associated with generative models. We therefore encourage responsible use of the proposed method and caution against deployment in contexts involving real identities, sensitive attributes, or high-stakes decision-making without additional safeguards.

\section{Reproducibility Statement}
\label{sec:code}
To ensure reproducibility, we provide detailed descriptions of the datasets, training setup, and hyperparameters in the main text and appendix. We will also release our code and checkpoints to further facilitate reproducibility soon.

%% file: tables/human_eval.tex
\begin{wraptable}{r}{0.5\textwidth}
\vspace{-1em}
\setlength{\belowcaptionskip}{0.2cm}
\caption{\textbf{Human evaluation results.}}
\label{tab:human_eval}
\resizebox{\hsize}{!}{
\renewcommand{\arraystretch}{1.50}  
\setlength\tabcolsep{3pt}  
\begin{tabular}{ccc}
\toprule
\textbf{Qwen-Image-Edit} & \textbf{LongCat-Image-Edit} & \textbf{UniCustom} \\ \midrule
0.291$^{\pm 0.015}$         & 0.142$^{\pm 0.018}$         & \textbf{0.567}$^{\pm 0.022}$ \\ \bottomrule     
\end{tabular}
}
\end{wraptable}

%% file: tables/cka.tex
\begin{table}[h]
\caption{\textbf{Layer-wise CKA analysis.}}
\label{tab:cka}
\renewcommand{\arraystretch}{1.1}
\setlength\tabcolsep{5pt}  
\centering
\resizebox{0.8\textwidth}{!}{
\begin{tabular}{lccccc}
\toprule
 & \textbf{Pre-VLM} & \textbf{Layer 7} & \textbf{Layer 14} & \textbf{Layer 21} & \textbf{Post-VLM} \\ \toprule
\textbf{Fused $\rightarrow$ VAE}                            & \textbf{0.541}   & \textbf{0.556}   & \textbf{0.684}    & \textbf{0.738}    & \textbf{0.938}    \\
ViT-only $\rightarrow$ VAE                                  & 0.363            & 0.364            & 0.369             & 0.358             & 0.351             \\ \midrule
\textbf{Fused $\rightarrow$ ViT}                            & \textbf{0.912}   & \textbf{0.892}   & \textbf{0.836}    & \textbf{0.828}    & \textbf{0.540}    \\
ViT-only $\rightarrow$ ViT                                  & 1.000            & 0.995            & 0.983             & 0.972             & 0.940             \\ \bottomrule
\end{tabular}
}
\end{table}

%% file: tables/probe.tex
\begin{table}[h]
\caption{\textbf{Layer-wise linear probing analysis.}}
\label{tab:probe}
\renewcommand{\arraystretch}{1.1}
\setlength\tabcolsep{5pt}  
\centering
\resizebox{0.7\textwidth}{!}{
\begin{tabular}{lccccc}
\toprule
 & \textbf{Pre-VLM} & \textbf{Layer 7} & \textbf{Layer 14} & \textbf{Layer 21} & \textbf{Post-VLM} \\ \toprule
\textbf{Fused}                     & \textbf{27.86}   & \textbf{27.50}   & \textbf{27.65}    & \textbf{27.74}    & \textbf{29.00}    \\
ViT-only                           & 12.45            & 12.47            & 12.35             & 12.09             & 11.87             \\ \bottomrule
\end{tabular}
}
\end{table}

%% file: bibfile.bib
@article{wu2025omnigen2,
  title={Omnigen2: Exploration to advanced multimodal generation},
  author={Wu, Chenyuan and Zheng, Pengfei and Yan, Ruiran and Xiao, Shitao and Luo, Xin and Wang, Yueze and Li, Wanli and Jiang, Xiyan and Liu, Yexin and Zhou, Junjie and others},
  journal={arXiv preprint arXiv:2506.18871},
  year={2025}
}

@article{wu2025qwen,
  title={Qwen-image technical report},
  author={Wu, Chenfei and Li, Jiahao and Zhou, Jingren and Lin, Junyang and Gao, Kaiyuan and Yan, Kun and Yin, Sheng-ming and Bai, Shuai and Xu, Xiao and Chen, Yilei and others},
  journal={arXiv preprint arXiv:2508.02324},
  year={2025}
}

@article{team2025longcat,
  title={Longcat-image technical report},
  author={Team, Meituan LongCat and Ma, Hanghang and Tan, Haoxian and Huang, Jiale and Wu, Junqiang and He, Jun-Yan and Gao, Lishuai and Xiao, Songlin and Wei, Xiaoming and Ma, Xiaoqi and others},
  journal={arXiv preprint arXiv:2512.07584},
  year={2025}
}

@article{bai2025qwen25vl,
      title={Qwen2.5-VL Technical Report}, 
      author={Shuai Bai and Keqin Chen and Xuejing Liu and Jialin Wang and Wenbin Ge and Sibo Song and Kai Dang and Peng Wang and Shijie Wang and Jun Tang and Humen Zhong and Yuanzhi Zhu and Mingkun Yang and Zhaohai Li and Jianqiang Wan and Pengfei Wang and Wei Ding and Zheren Fu and Yiheng Xu and Jiabo Ye and Xi Zhang and Tianbao Xie and Zesen Cheng and Hang Zhang and Zhibo Yang and Haiyang Xu and Junyang Lin},
      journal={arXiv preprint arXiv:2502.13923},
      year={2025}
}

@inproceedings{lin2014microsoft,
  title={Microsoft coco: Common objects in context},
  author={Lin, Tsung-Yi and Maire, Michael and Belongie, Serge and Hays, James and Perona, Pietro and Ramanan, Deva and Doll{\'a}r, Piotr and Zitnick, C Lawrence},
  booktitle={European conference on computer vision},
  pages={740--755},
  year={2014},
  organization={Springer}
}

@article{ye2025echo,
  title={Echo-4o: Harnessing the power of gpt-4o synthetic images for improved image generation},
  author={Ye, Junyan and Jiang, Dongzhi and Wang, Zihao and Zhu, Leqi and Hu, Zhenghao and Huang, Zilong and He, Jun and Yan, Zhiyuan and Yu, Jinghua and Li, Hongsheng and others},
  journal={arXiv preprint arXiv:2508.09987},
  year={2025}
}

@article{yang2023hq,
  title={Hq-50k: A large-scale, high-quality dataset for image restoration},
  author={Yang, Qinhong and Chen, Dongdong and Tan, Zhentao and Liu, Qiankun and Chu, Qi and Bao, Jianmin and Yuan, Lu and Hua, Gang and Yu, Nenghai},
  journal={arXiv preprint arXiv:2306.05390},
  year={2023}
}

@inproceedings{ju2023human,
  title={Human-art: A versatile human-centric dataset bridging natural and artificial scenes},
  author={Ju, Xuan and Zeng, Ailing and Wang, Jianan and Xu, Qiang and Zhang, Lei},
  booktitle={Proceedings of the IEEE/CVF conference on computer vision and pattern recognition},
  pages={618--629},
  year={2023}
}

@inproceedings{
xu2026withanyone,
title={WithAnyone: Toward Controllable and {ID} Consistent Image Generation},
author={Hengyuan Xu and Wei Cheng and Peng Xing and Yixiao Fang and Shuhan Wu and Rui Wang and Xianfang Zeng and Daxin Jiang and Gang YU and Xingjun Ma and Yu-Gang Jiang},
booktitle={The Fourteenth International Conference on Learning Representations},
year={2026}
}

@inproceedings{karras2019style,
  title={A style-based generator architecture for generative adversarial networks},
  author={Karras, Tero and Laine, Samuli and Aila, Timo},
  booktitle={Proceedings of the IEEE/CVF conference on computer vision and pattern recognition},
  pages={4401--4410},
  year={2019}
}

@misc{unsplash_datasets_github,
  title        = {The Unsplash Dataset},
  author       = {{Unsplash}},
  year         = {2025},
  howpublished = {\url{https://github.com/unsplash/datasets}}
}

@article{zhang2025bee,
  title={Bee: A High-Quality Corpus and Full-Stack Suite to Unlock Advanced Fully Open MLLMs},
  author={Zhang, Yi and Ni, Bolin and Chen, Xin-Sheng and Zhang, Heng-Rui and Rao, Yongming and Peng, Houwen and Lu, Qinglin and Hu, Han and Guo, Meng-Hao and Hu, Shi-Min},
  journal={arXiv preprint arXiv:2510.13795},
  year={2025}
}

@article{wei2025mico,
  title={MICo-150K: A Comprehensive Dataset Advancing Multi-Image Composition},
  author={Wei, Xinyu and Cen, Kangrui and Wei, Hongyang and Guo, Zhen and Li, Bairui and Wang, Zeqing and Zhang, Jinrui and Zhang, Lei},
  journal={arXiv preprint arXiv:2512.07348},
  year={2025}
}

@article{lin2025uniworld,
  title={Uniworld-v1: High-resolution semantic encoders for unified visual understanding and generation},
  author={Lin, Bin and Li, Zongjian and Cheng, Xinhua and Niu, Yuwei and Ye, Yang and He, Xianyi and Yuan, Shenghai and Yu, Wangbo and Wang, Shaodong and Ge, Yunyang and others},
  journal={arXiv preprint arXiv:2506.03147},
  year={2025}
}

@article{ye2025imgedit,
  title={Imgedit: A unified image editing dataset and benchmark},
  author={Ye, Yang and He, Xianyi and Li, Zongjian and Lin, Bin and Yuan, Shenghai and Yan, Zhiyuan and Hou, Bohan and Yuan, Li},
  journal={arXiv preprint arXiv:2505.20275},
  year={2025}
}

@inproceedings{kuprashevich2026nohumansrequired,
  title={Nohumansrequired: Autonomous high-quality image editing triplet mining},
  author={Kuprashevich, Maksim and Alekseenko, Grigorii and Tolstykh, Irina and Fedorov, Georgii and Suleimanov, Bulat and Dokholyan, Vladimir and Gordeev, Aleksandr},
  booktitle={Proceedings of the IEEE/CVF Winter Conference on Applications of Computer Vision},
  pages={6059--6068},
  year={2026}
}

@article{chen2025blip3,
  title={Blip3-o: A family of fully open unified multimodal models-architecture, training and dataset},
  author={Chen, Jiuhai and Xu, Zhiyang and Pan, Xichen and Hu, Yushi and Qin, Can and Goldstein, Tom and Huang, Lifu and Zhou, Tianyi and Xie, Saining and Savarese, Silvio and others},
  journal={arXiv preprint arXiv:2505.09568},
  year={2025}
}

@article{lin2024open,
  title={Open-sora plan: Open-source large video generation model},
  author={Lin, Bin and Ge, Yunyang and Cheng, Xinhua and Li, Zongjian and Zhu, Bin and Wang, Shaodong and He, Xianyi and Ye, Yang and Yuan, Shenghai and Chen, Liuhan and others},
  journal={arXiv preprint arXiv:2412.00131},
  year={2024}
}

@inproceedings{
wang2025msdiffusion,
title={{MS}-Diffusion: Multi-subject Zero-shot Image Personalization with Layout Guidance},
author={Xierui Wang and Siming Fu and Qihan Huang and Wanggui He and Hao Jiang},
booktitle={The Thirteenth International Conference on Learning Representations},
year={2025}
}

@article{ye2023ip,
  title={Ip-adapter: Text compatible image prompt adapter for text-to-image diffusion models},
  author={Ye, Hu and Zhang, Jun and Liu, Sibo and Han, Xiao and Yang, Wei},
  journal={arXiv preprint arXiv:2308.06721},
  year={2023}
}

@inproceedings{
li2023blipdiffusion,
title={{BLIP}-Diffusion: Pre-trained Subject Representation for Controllable Text-to-Image Generation and Editing},
author={Dongxu Li and Junnan Li and Steven Hoi},
booktitle={Thirty-seventh Conference on Neural Information Processing Systems},
year={2023},
}

@article{garibi2025tokenverse,
  title={Tokenverse: Versatile multi-concept personalization in token modulation space},
  author={Garibi, Daniel and Yadin, Shahar and Paiss, Roni and Tov, Omer and Zada, Shiran and Ephrat, Ariel and Michaeli, Tomer and Mosseri, Inbar and Dekel, Tali},
  journal={ACM Transactions On Graphics (TOG)},
  volume={44},
  number={4},
  pages={1--11},
  year={2025},
  publisher={ACM New York, NY, USA}
}

@inproceedings{tan2025ominicontrol,
  title={Ominicontrol: Minimal and universal control for diffusion transformer},
  author={Tan, Zhenxiong and Liu, Songhua and Yang, Xingyi and Xue, Qiaochu and Wang, Xinchao},
  booktitle={2025 IEEE/CVF International Conference on Computer Vision (ICCV)},
  pages={14940--14950},
  year={2025},
  organization={IEEE}
}

@inproceedings{wu2025less,
  title={Less-to-more generalization: Unlocking more controllability by in-context generation},
  author={Wu, Shaojin and Huang, Mengqi and Wu, Wenxu and Cheng, Yufeng and Ding, Fei and He, Qian},
  booktitle={Proceedings of the IEEE/CVF International Conference on Computer Vision},
  pages={18682--18692},
  year={2025}
}

@inproceedings{
she2026mosaic,
title={{MOSAIC}: Multi-Subject Personalized Generation via Correspondence-Aware Alignment and Disentanglement},
author={Dong She and Siming Fu and Mushui Liu and Qiaoqiao Jin and Hualiang Wang and Mu Liu and Jidong Jiang},
booktitle={The Fourteenth International Conference on Learning Representations},
year={2026}
}

@inproceedings{mou2025dreamo,
  title={Dreamo: A unified framework for image customization},
  author={Mou, Chong and Wu, Yanze and Wu, Wenxu and Guo, Zinan and Zhang, Pengze and Cheng, Yufeng and Luo, Yiming and Ding, Fei and Zhang, Shiwen and Li, Xinghui and others},
  booktitle={Proceedings of the SIGGRAPH Asia 2025 Conference Papers},
  pages={1--12},
  year={2025}
}

@article{chen2026xverse,
  title={Xverse: Consistent multi-subject control of identity and semantic attributes via dit modulation},
  author={Chen, Bowen and Sun, Haomiao and Chen, Li and Wang, Xu and Du, Daniel and Wu, Xinglong and others},
  journal={Advances in Neural Information Processing Systems},
  volume={38},
  pages={7437--7459},
  year={2026}
}

@article{cheng2025umo,
  title={UMO: Scaling Multi-Identity Consistency for Image Customization via Matching Reward},
  author={Cheng, Yufeng and Wu, Wenxu and Wu, Shaojin and Huang, Mengqi and Ding, Fei and He, Qian},
  journal={arXiv preprint arXiv:2509.06818},
  year={2025}
}

@article{wang2025psr,
  title={PSR: Scaling Multi-Subject Personalized Image Generation with Pairwise Subject-Consistency Rewards},
  author={Wang, Shulei and Wei, Longhui and He, Xin and Ouyang, Jianbo and Lu, Hui and Zhao, Zhou and Tian, Qi},
  journal={arXiv preprint arXiv:2512.01236},
  year={2025}
}

@article{dalva2025canvas,
  title={Canvas-to-Image: Compositional Image Generation with Multimodal Controls},
  author={Dalva, Yusuf and Qian, Guocheng Gordon and Goldenberg, Maya and Chen, Tsai-Shien and Aberman, Kfir and Tulyakov, Sergey and Yanardag, Pinar and Wang, Kuan-Chieh Jackson},
  journal={arXiv preprint arXiv:2511.21691},
  year={2025}
}

@inproceedings{
pan2024kosmosg,
title={Kosmos-G: Generating Images in Context with Multimodal Large Language Models},
author={Xichen Pan and Li Dong and Shaohan Huang and Zhiliang Peng and Wenhu Chen and Furu Wei},
booktitle={The Twelfth International Conference on Learning Representations},
year={2024}
}

@article{liu2025step1x,
  title={Step1x-edit: A practical framework for general image editing},
  author={Liu, Shiyu and Han, Yucheng and Xing, Peng and Yin, Fukun and Wang, Rui and Cheng, Wei and Liao, Jiaqi and Wang, Yingming and Fu, Honghao and Han, Chunrui and others},
  journal={arXiv preprint arXiv:2504.17761},
  year={2025}
}

@inproceedings{
fu2024guiding,
title={Guiding Instruction-based Image Editing via Multimodal Large Language Models},
author={Tsu-Jui Fu and Wenze Hu and Xianzhi Du and William Yang Wang and Yinfei Yang and Zhe Gan},
booktitle={The Twelfth International Conference on Learning Representations},
year={2024},
}

@article{cai2025z,
  title={Z-image: An efficient image generation foundation model with single-stream diffusion transformer},
  author={Cai, Huanqia and Cao, Sihan and Du, Ruoyi and Gao, Peng and Hoi, Steven and Hou, Zhaohui and Huang, Shijie and Jiang, Dengyang and Jin, Xin and Li, Liangchen and others},
  journal={arXiv preprint arXiv:2511.22699},
  year={2025}
}

@misc{gpt-image-2,
  title        = {Introducing ChatGPT Images 2.0},
  author       = {OpenAI},
  year         = {2026},
  howpublished = {\url{https://openai.com/index/introducing-chatgpt-images-2-0/}}
}

@misc{nanobanan,
    title = {Nano Banana 2: Combining Pro capabilities with lightning-fast speed},
    author = {Google},
    year = {2026},
    howpublished = {\url{https://blog.google/innovation-and-ai/technology/ai/nano-banana-2/}}
}

@article{wang2025scone,
  title={Scone: Bridging Composition and Distinction in Subject-Driven Image Generation via Unified Understanding-Generation Modeling},
  author={Wang, Yuran and Zeng, Bohan and Tong, Chengzhuo and Liu, Wenxuan and Shi, Yang and Ma, Xiaochen and Liang, Hao and Zhang, Yuanxing and Zhang, Wentao},
  journal={arXiv preprint arXiv:2512.12675},
  year={2025}
}

@article{deng2025cinema,
  title={Cinema: Coherent multi-subject video generation via mllm-based guidance},
  author={Deng, Yufan and Guo, Xun and Wang, Yizhi and Fang, Jacob Zhiyuan and Wang, Angtian and Yuan, Shenghai and Yang, Yiding and Liu, Bo and Huang, Haibin and Ma, Chongyang},
  journal={arXiv preprint arXiv:2503.10391},
  year={2025}
}

@article{hu2025hunyuancustom,
  title={Hunyuancustom: A multimodal-driven architecture for customized video generation},
  author={Hu, Teng and Yu, Zhentao and Zhou, Zhengguang and Liang, Sen and Zhou, Yuan and Lin, Qin and Lu, Qinglin},
  journal={arXiv preprint arXiv:2505.04512},
  year={2025}
}

@inproceedings{hore2010image,
  title={Image quality metrics: PSNR vs. SSIM},
  author={Hore, Alain and Ziou, Djemel},
  booktitle={2010 20th international conference on pattern recognition},
  pages={2366--2369},
  year={2010},
  organization={IEEE}
}

@article{labs2025flux,
  title={FLUX. 1 Kontext: Flow Matching for In-Context Image Generation and Editing in Latent Space},
  author={Labs, Black Forest and Batifol, Stephen and Blattmann, Andreas and Boesel, Frederic and Consul, Saksham and Diagne, Cyril and Dockhorn, Tim and English, Jack and English, Zion and Esser, Patrick and others},
  journal={arXiv preprint arXiv:2506.15742},
  year={2025}
}

@article{wu2025uso,
  title={Uso: Unified style and subject-driven generation via disentangled and reward learning},
  author={Wu, Shaojin and Huang, Mengqi and Cheng, Yufeng and Wu, Wenxu and Tian, Jiahe and Luo, Yiming and Ding, Fei and He, Qian},
  journal={arXiv preprint arXiv:2508.18966},
  year={2025}
}

@article{deng2025emerging,
  title={Emerging properties in unified multimodal pretraining},
  author={Deng, Chaorui and Zhu, Deyao and Li, Kunchang and Gou, Chenhui and Li, Feng and Wang, Zeyu and Zhong, Shu and Yu, Weihao and Nie, Xiaonan and Song, Ziang and others},
  journal={arXiv preprint arXiv:2505.14683},
  year={2025}
}

@article{qian2025pico,
  title={Pico-banana-400k: A large-scale dataset for text-guided image editing},
  author={Qian, Yusu and Bocek-Rivele, Eli and Song, Liangchen and Tong, Jialing and Yang, Yinfei and Lu, Jiasen and Hu, Wenze and Gan, Zhe},
  journal={arXiv preprint arXiv:2510.19808},
  year={2025}
}

@inproceedings{wang2025precise,
  title={Precise, fast, and low-cost concept erasure in value space: Orthogonal complement matters},
  author={Wang, Yuan and Li, Ouxiang and Mu, Tingting and Hao, Yanbin and Liu, Kuien and Wang, Xiang and He, Xiangnan},
  booktitle={2025 IEEE/CVF Conference on Computer Vision and Pattern Recognition (CVPR)},
  pages={28759--28768},
  year={2025},
  organization={IEEE}
}

@inproceedings{qiu2025accelerating1,
  title={Accelerating diffusion transformer via gradient-optimized cache},
  author={Qiu, Junxiang and Liu, Lin and Wang, Shuo and Lu, Jinda and Chen, Kezhou and Hao, Yanbin},
  booktitle={2025 IEEE/CVF International Conference on Computer Vision (ICCV)},
  pages={17608--17617},
  year={2025},
  organization={IEEE}
}

@inproceedings{qiu2025accelerating2,
  title={Accelerating diffusion transformer via error-optimized cache},
  author={Qiu, Junxiang and Wang, Shuo and Lu, Jinda and Liu, Lin and Jiang, Houcheng and Zhu, Xingyu and Hao, Yanbin},
  booktitle={Proceedings of the 33rd ACM International Conference on Multimedia},
  pages={9588--9597},
  year={2025}
}

@inproceedings{li2026easier,
  title={Easier Painting Than Thinking: Can Text-to-Image Models Set the Stage, but Not Direct the Play?},
  author={Li, Ouxiang and Wang, Yuan and Hu, Xinting and Huang, Huijuan and Chen, Rui and Ou, Jiarong and Tao, Xin and Wan, Pengfei and Qi, Xiaojuan and Feng, Fuli},
  booktitle={International Conference on Learning Representations},
  volume={2026},
  pages={86729--86758},
  year={2026}
}

@inproceedings{li2026speed,
  title={Speed: Scalable, precise, and efficient concept erasure for diffusion models},
  author={Li, Ouxiang and Wang, Yuan and Hu, Xinting and Jiang, Houcheng and Hao, Yanbin and Ma, James and Feng, Fuli and others},
  booktitle={International Conference on Learning Representations},
  volume={2026},
  pages={66777--66803},
  year={2026}
}
